\documentclass[journal]{IEEEtran}
\usepackage{amssymb}
\usepackage{times}
\usepackage{graphicx}
\usepackage{subfigure}
\usepackage{multirow}
\usepackage{multicol}
\usepackage{amsmath}
\usepackage{bm}
\allowdisplaybreaks

\usepackage{dashrule}
\usepackage{algorithm}
\usepackage{algorithmic}
\usepackage[table*]{xcolor}
\usepackage{threeparttable}
\usepackage{dcolumn}
\usepackage{multirow}
\usepackage{multicol}
\usepackage{booktabs}
\usepackage{subfigure}
\usepackage{hyperref}  

\usepackage{amsthm}

\usepackage{latexsym}

\usepackage{rotating}
\usepackage{textcomp}
\usepackage{cite}
\usepackage{mathrsfs}
\usepackage{colortbl}
\definecolor{light-gray}{gray}{0.75} 

\ifCLASSINFOpdf

\else

\fi

\begin{document}
	%
	\title{Language Model Evolutionary Algorithms \\ for Recommender Systems: \\ Benchmarks and Algorithm Comparisons}
	%
	%
	%
	
	\author{Jiao Liu,
		Zhu Sun,
		Shanshan Feng,
		Caishun Chen,
		and Yew-Soon Ong
		\thanks{This research is partially supported by the Ministry of Education, Singapore, under its MOE AcRF Tier 1, SUTD Kickstarter Initiative (SKI 2021\_06\_12), partially supported by MTI under its AI Centre of Excellence for Manufacturing (AIMfg) (Award W25MCMF014), and the College of Computing and Data Science, Nanyang Technological University. Moreover, the authors acknowledge the use of GPT-4o to identify improvements in the writing, prompts used with ``Please refine and optimize the description of the following, and make it be suitable for a research paper". (\textit{Corresponding author: Zhu Sun})}
		\thanks{Jiao Liu is with the College of Computing \& Data Science, Nanyang Technological University, Singapore (e-mail: jiao.liu@ntu.edu.sg).}
		\thanks{Zhu Sun is with the Information Systems Technology and Design, Singapore University of Technology and Design, Singapore (e-mail: zhu\_sun@sutd.edu.sg).}
		\thanks{Shanshan Feng is with the Centre for Frontier AI Research, A*STAR and the Institute of High Performance Computing, A*STAR, Singapore (e-mail: victor\_fengss@foxmail.com)}
		\thanks{Caishun Chen is with the Centre for Frontier AI Research, A*STAR and the Institute of High Performance Computing, A*STAR, Singapore (e-mail: Chen\_Caishun@cfar.a-star.edu.sg)}
		\thanks{Yew-Soon Ong is with the College of Computing \& Data Science, Nanyang Technological University, Singapore, and also with the Centre for Frontier AI Research, A*STAR and the Institute of High Performance Computing, A*STAR, Singapore (e-mail: asysong@ntu.edu.sg; ong\_yew\_soon@a-star.edu.sg).}
	}
	
	%
	%

	\markboth{Journal of \LaTeX\ Class Files,~Vol.~14, No.~8, August~2015}%
	{Shell \MakeLowercase{\textit{et al.}}: Bare Demo of IEEEtran.cls for IEEE Journals}
	%



	\maketitle
	
	\begin{abstract}
		In the evolutionary computing community, the remarkable language-handling capabilities and reasoning power of large language models (LLMs) have significantly enhanced the functionality of evolutionary algorithms (EAs), enabling them to tackle optimization problems involving structured language or program code. Although this field is still in its early stages, its impressive potential has led to the development of various LLM-based EAs. To effectively evaluate the performance and practical applicability of these LLM-based EAs, benchmarks with real-world relevance are essential. In this paper, we focus on LLM-based recommender systems (RSs) and introduce a benchmark problem set, named RSBench, specifically designed to assess the performance of LLM-based EAs in recommendation prompt optimization. RSBench emphasizes session-based recommendations, aiming to discover a set of Pareto optimal prompts that guide the recommendation process, providing accurate, diverse, and fair recommendations. We develop three LLM-based EAs based on established EA frameworks and experimentally evaluate their performance using RSBench. Our study offers valuable insights into the application of EAs in LLM-based RSs. Additionally, we explore key components that may influence the overall performance of the RS, providing meaningful guidance for future research on the development of LLM-based EAs in RSs. The source code of the proposed RSBench can be found at https://github.com/LiuJ-2023/RSBench/tree/main.
	\end{abstract}
	
	
	%
	\IEEEpeerreviewmaketitle

	\section{Introduction}
	Evolutionary algorithms (EAs), inspired by natural selection and genetic principles, constitute a powerful class of optimization techniques~\cite{coello2007evolutionary}. Unlike traditional methods, EAs do not rely on gradient information, making them highly suitable for tackling complex black-box optimization problems that commonly arise in scientific~\cite{oganov2019structure} and engineering~\cite{deb1999evolutionary} fields. Moreover, EAs are particularly effective for multiobjective optimization problems, where multiple conflicting objectives require simultaneous optimization~\cite{branke2008multiobjective}. Their population-based search strategy, which leverages implicit parallelism, allows EAs to explore diverse solution spaces within a single optimization run. This approach yields a set of optimized solutions—referred to as nondominated or Pareto optimal solutions—that align with the Pareto front (PF), representing the best achievable performance trade-offs in the objective space. {Meanwhile, diversity-keeping mechanisms within EAs facilitate the identification of solutions that are well-distributed along the PF, resulting in a more accurate approximation of the PF.}~\cite{9005525}.
	
	{Large language models (LLMs)\cite{zhao2023survey,OpenAI} have significantly advanced language processing technology}, exhibiting exceptional performance in natural language processing tasks and fostering transformative applications across fields such as mechanical engineering~\cite{buehler2024mechgpt}, management~\cite{li2023large}, and chemistry~\cite{bran2023chemcrow}. This advancement has created promising opportunities to enhance EAs~\cite{wu2024evolutionary}. On one hand, the black-box optimization capabilities of EAs are well-suited for LLM-based applications. Due to the high computational demands of LLM training and inference, deployment is largely restricted to corporations like OpenAI, Google, and Baidu, which often limit access to model parameters for security and privacy, providing only API-based querying~\cite{wu2024evolutionary}. Consequently, users often treat LLMs as black-box systems, and EAs, as classic and effective black-box optimization algorithms, present a promising approach for refining LLM applications~\cite{guo2023connecting}. On the other hand, the generative capabilities of LLMs can help expand the functional scope of traditional EAs, which typically represent solutions as low-dimensional continuous or discrete vectors~\cite{10.1145/3694791}. Integrating LLMs enables EAs to process language and code, allowing them to tackle complex tasks such as prompt optimization~\cite{xu2022gps}, neural architecture search~\cite{nasir2024llmatic}, and code generation~\cite{bradley2024openelm}. Consequently, recent research has increasingly explored hybrid approaches that combine EAs with LLMs, applying these techniques to both traditional optimization challenges and novel application domains.
	
	Recommender systems (RSs) play an increasingly crucial role on modern online platforms~\cite{bobadilla2013recommender}, and the emergence of LLMs also has introduced new opportunities for their development and enhancement~\cite{10506571}. 
	{
		Traditional RS algorithms, such as collaborative filtering~\cite{koren2021advances}, primarily rely on ID-based information while overlooking the rich textual data generated through user interactions on the platform~\cite{ren2024representation}. In fact, leveraging these textual data can help mitigate inherent limitations in traditional recommendation algorithms, such as the cold-start problem in collaborative filtering~\cite{ji2024genrec}, while also capturing users' current preferences and intentions~\cite{10.1145/3626772.3657688}. The powerful text-processing capabilities of LLMs further enable RSs to effectively utilize textual data, leading to more sophisticated and context-aware recommendations.
	}
	In the early stages, language models such as BERT~\cite{kenton2019bert} were used to embed language elements from the recommendation process into suitable vectors, supporting the training of more accurate recommendation models~\cite{hoang2021using}. As LLMs evolved, pre-training~\cite{wu2020ptum} and fine-tuning~\cite{friedman2023leveraging} paradigms were developed, leading to the creation of even more powerful recommendation performance~\cite{mao2023unitrec}. Recently, with the advent of models like ChatGPT~\cite{OpenAI}, the advanced language processing capabilities of LLMs enable the creation of robust RSs through simple prompts~\cite{wang2023recmind, huang2023recommender, 10.1145/3626772.3657688} or in-context learning~\cite{liu2023chatgpt, zhiyuli2023bookgpt}. 
	
	In prompt-based LLM-driven RSs, the design of prompts plays a pivotal role in shaping system performance~\cite{10.1145/3626772.3657688}. As a result, finding the optimal prompt becomes a critical priority when developing such systems. Additionally, in RSs, it is essential for recommendations not only to align with user-specific interests but also to consider factors such as diversity and fairness to enhance overall user satisfaction~\cite{zhang2008avoiding,li2022fairness}. Multiobjective optimization serves as an ideal theoretical framework for this purpose~\cite{rodriguez2012multiple}. By leveraging this approach, the system can provide users with a set of Pareto-optimal solutions that balance criteria like recommendation accuracy, diversity, and fairness, thus allowing users to select options that best meet their personalized preferences. Whether addressing black-box prompt optimization or multiobjective optimization, EAs have demonstrated remarkable capabilities. Consequently, EAs hold significant potential for application in LLM-based RSs. {However, to date, there are almost no benchmark problems specifically addressing the application of LLM-based EAs to LLM-based RSs.} To bridge this gap, this paper explores the application of language model based EAs in RSs. The key contributions of this paper are mainly focus on the follow aspects:
	\begin{itemize}
		\item {\color{blue}In this work, we introduce RSBench, a benchmark specifically designed for LLM-based RSs. RSBench provides a convenient and accessible testbed that enables EA researchers to explore, validate, and advance evolutionary algorithms in the context of LLM-based recommendation. By bridging this gap, RSBench encourages stronger engagement between the EA and RS communities and promotes deeper investigation into the potential and practical value of EA for real-world RS challenges. To support reproducibility and further research, we release the accompanying code at https://github.com/LiuJ-2023/RSBench, serving as a foundational resource for understanding and advancing this field.}
				
		\item {LLM-based initialization and offspring generation are designed and incorporated into three representative multiobjective EAs—\, i.e., NSGA-II, MOEA/D, and IBEA—to support their application in solving RSBench problems.} Their performance is then evaluated experimentally, providing valuable insights and guidance for the application of LLM-based EAs in LLM-based RSs. 
		\item  We further investigate key components that may impact the performance of the final RS, including the effectiveness of the LLM-based initialization and the LLM-based offspring generation with the supporting of the crossover and mutation operators. These investigations are {expected to point out challenges and potential research questions for future research on LLM-based EAs in RSs}.
		\item {\color{blue}RSBench can also serve as a test function suite for prompt optimization in application domains. It incorporates characteristics such as expansionality, uncertainty, and language-space search, thereby reflecting many of the distinctive properties of prompt optimization problems at the LLM application level.}
	\end{itemize}
	
	
	The remainder of this paper is organized as follows: Section II provides a review of related work and background concepts. Section III introduces RSBench. In Section IV, we present the three LLM-based multiobjective EAs, based on the classical NSGA-II, MOEA/D, and IBEA frameworks. Section V discusses the systematic experimental investigations, {which aims to provide insights and guidance for researchers in the related area.} In Section VI, we explore promising research directions based on RSBench and highlight pressing issues for future exploration. Finally, Section VII concludes the paper.
	
	\section{Preliminaries}
	
	\subsection{Evolutionary Multiobjective Optimization}
	Without loss of generality, a multiobjective optimization problem can be stated as:
	\begin{equation}\label{Eq:Layout_Problem}
	\begin{aligned}
	\max: \ & \textbf{F}(\textbf{p}) = \{F_1(\textbf{p}),\ldots,F_m(\textbf{p})\} \\
	\text{s.t.}\ & \textbf{p} \in \mathcal{X} \\
	\end{aligned}
	\end{equation}
	where $F_i(\textbf{p}),(i \in \{1,\ldots,m\})$ is the $i$th objective function, $m$ is the number of objectives, $\textbf{p}$ is the decision variable which is considered as a prompt in this paper, $\mathcal{X}$ is the decision space (or search space). Key concepts associated with the formulation in \eqref{Eq:Layout_Problem} are as follows~\cite{branke2008multiobjective}:
	\begin{itemize}
		\item \textit{Pareto Dominance}: For decision vectors $\textbf{p}_a$ and $\textbf{p}_b$, if $\forall i \in \{1,2,\ldots,m\}$, $F_{i}(\textbf{p}_a) \geq F_{i}(\textbf{p}_b)$ and $\exists j \in \{1,2,\ldots,m\}$, $F_{j}(\textbf{p}_a) > F_{j}(\textbf{p}_b)$, $\textbf{p}_a$ is said to {Pareto dominate} $\textbf{p}_b$.
		
		\item \textit{Pareto Optimal Solution}: If no decision vector in $\mathcal{X}$ Pareto dominates $\textbf{p}_a$, then $\textbf{p}_a$ is a {Pareto optimal} solution.
		
		\item \textit{Pareto Set}: The set of all Pareto optimal solutions forms the Pareto set in decision space.
		
		\item \textit{Pareto Front}: The image of the Pareto set in the objective space forms the Pareto front.
	\end{itemize}
	
	EAs, with their inherent parallelism in population-based search, have long been instrumental in solving multiobjective optimization problems, as noted in \cite{qian-good}. Multiobjective EAs are typically classified into three main categories: domination-based methods~\cite{nsga2, spea2}, decomposition-based methods~\cite{decomposition-survey, moead}, and indicator-based methods~\cite{fvmoea, ibea}. Over recent decades, these three types of multiobjective EAs have consistently demonstrated strong performance across various multiobjective optimization tasks and have been extended to address many-objective optimization problems~\cite{yang2013grid}, constrained multiobjective optimization problems~\cite{8632683,8624421}, and multiobjective combinatorial optimization problems~\cite{shao2021multi}. Distinct from previous studies, this paper further explores the integration of LLMs within these classical EA frameworks, examining their potential applications in RSs.
	
	\subsection{Large Language Models in Evolutionary Computation}
	The integration of LLMs and evolutionary computation can be categorized into two primary directions~\cite{wu2024evolutionary}. The first direction focuses on utilizing LLMs to enhance EAs, thereby directly addressing optimization challenges or achieving improved performance on various mathematical programming problems~\cite{liu2024large}. These methods are founded on the premise that the extensive knowledge acquired by LLMs can guide EAs in developing more effective and efficient optimization frameworks. One of the earliest works in this domain is OPRO~\cite{yang2024largelanguagemodelsoptimizers}, which adapts LLMs to solve optimization problems by generating promising solutions for combinatorial optimization challenges, such as the traveling salesman problem. Liu \textit{et al.}~\cite{liu2024large} propose LMEA, which leverages LLMs for crossover and mutation operations and constructs prompts in each generation to assist the LLM in selecting parent solutions from the current population. In addition to these methods, there is ongoing research investigating the effectiveness of various LLM-based components, including crossover/mutation~\cite{10.1145/3694791}, surrogate models~\cite{hao2024large}, and their extension to constrained optimization~\cite{wang2024large}, multiobjective optimization~\cite{liu2023large} and/or multitask optimization~\cite{wong2024llm2fea}.
	
	The second direction involves leveraging the powerful black-box optimization capabilities of EAs to address optimization problems within LLM-based application scenarios~\cite{hemberg2024evolving, brownlee2023enhancing, yu2023gpt}. One prominent application area for EAs is prompt engineering~\cite{diao2022black}, where various studies have explored the use of EAs to identify optimal prompts, represented as both textual instructions and embedded vectors, to guide LLMs in executing complex language tasks~\cite{saletta2024exploring, guo2023connecting, xu2023wizardlm}. Another significant application is code generation, which benefits from the robust code generation capabilities of LLMs alongside the optimization strengths of EAs to produce high-quality code~\cite{li2022competition}. Recently, several EA-LLM-based methods for code generation have been proposed~\cite{brownlee2023enhancing, wu2023deceptprompt}. Additionally, with the support of LLMs' reasoning and coding abilities, EAs have also been effectively utilized for neural network architecture search, yielding impressive results~\cite{yu2023gpt, morris2024llm}.
	
	Despite the increasing number of studies combining EAs and LLMs to achieve various objectives, to the best of our knowledge, no research has yet applied these methodologies to RSs. In contrast to prior works, this paper primarily focuses on the application of EAs in the context of LLM-based RSs.
	
	\subsection{Recommender Systems in Evolutionary Computation}
	EAs are well-suited for RSs due to their ability to handle multiobjective optimization, aligning with the multi-metric nature of recommendation tasks. Geng \textit{et al.}\cite{geng2015nnia} framed recommendation as a multiobjective optimization problem, proposing NNIA-RS, which utilizes a nondominated neighbor immune algorithm as an optimizer. Similarly, Cui \textit{et al.}\cite{cui2017novel} applied multiobjective EAs to balance precision and diversity in recommendations. Beyond these approaches, recent advances, including many-objective optimization~\cite{cai2020hybrid}, evolutionary clustering~\cite{rana2014evolutionary}, and hybrid methods combining EAs with neural networks~\cite{asgarnezhad2022effective}, have further advanced EA applications in RSs. However, these methods are limited by their reliance on small-scale user-item interactions and struggle to integrate session-based natural language features, constraining their scalability and applicability in real-world contexts~\cite{milojkovic2019multi}.
	
	Recently, LLM-based RSs have demonstrated strong potential for generating context-aware, personalized recommendations, often requiring multiobjective optimization to incorporate human-centered, customized preferences~\cite{10506571}. Nevertheless, limited studies have explored the application of multiobjective EAs within LLM-based recommendation frameworks.
	
	\section{RSBench: The Proposed Benchmark}
	
	\subsection{Description of the Recommendation Task}\label{sec:task_description}
	\begin{figure}[!t]
		\begin{center}
			\includegraphics[width=1\columnwidth]{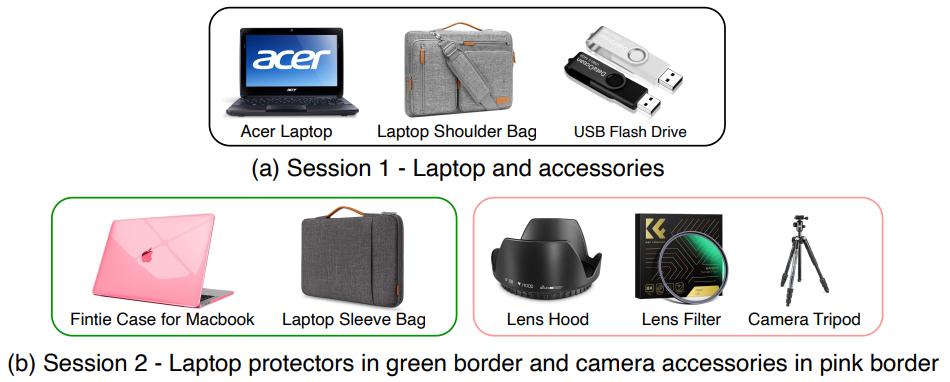}
			\caption{Examples of user sessions with various intents~\cite{10.1145/3626772.3657688}.}
			\label{fig:rs_intents}
		\end{center}
	\end{figure}
	
	\begin{figure}[!t]
		\begin{center}
			\includegraphics[width=1\columnwidth]{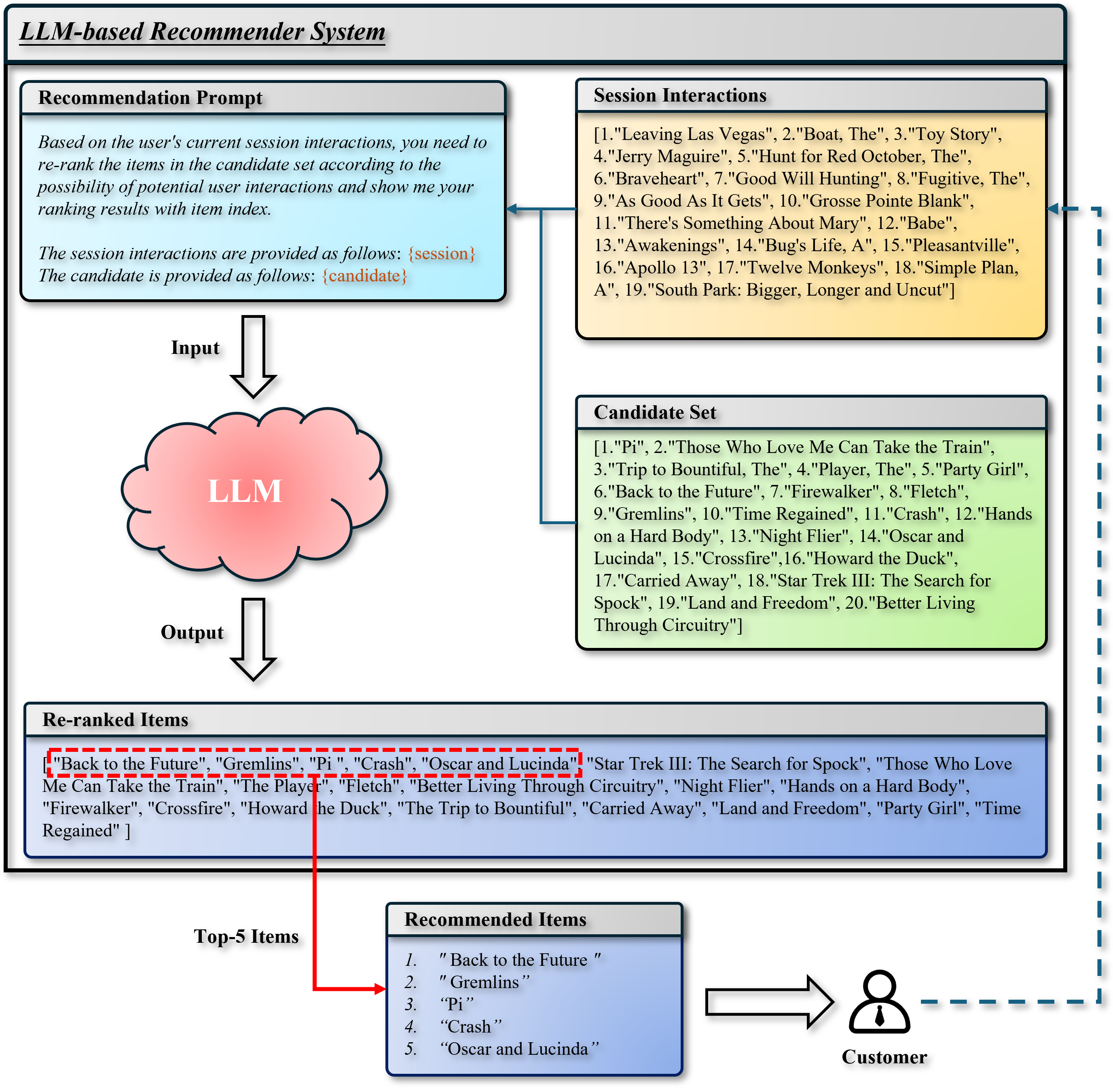}
			\vspace{-2mm}
			\caption{The example of the LLM-based RS.}
			\label{fig:rs_example}
		\end{center}
	\end{figure}
	
	In this paper, we focus on session-based recommendation~\cite{10.1145/3626772.3657688}, which aims to predict the next item a user will interact with based on short, anonymous behavior sessions. This approach is grounded in the observation that different sessions may reveal distinct user intents. In Fig.~\ref{fig:rs_intents}, we illustrate two real sessions from the Amazon Electronics dataset~\cite{ni2019justifying}. The first session, shown in (a), reflects a single primary intent—shopping for a laptop and related accessories—whereas the second session, shown in (b), demonstrates two major intents: shopping for laptop protectors and camera accessories. A key challenge in session-based recommendation is accurately inferring the user's intent based on their interactions to provide relevant product suggestions. Leveraging the powerful reasoning capabilities of LLMs, the RS has the potential to more effectively capture latent user intents within a session, thereby improving recommendation performance~\cite{10.1145/3626772.3657688}. This paper focuses on the development and evaluation of an LLM-assisted session-based RS.
	
	The RS considered in this paper consists of three key components:
	\begin{itemize} 
		
		\item \textit{Recommendation Prompt}: The recommendation prompt describes the current task, guiding the LLM to generate appropriate recommendations for the customer.
		
		\item \textit{Session Interactions}: For each customer, a set of session interactions is provided to the LLM. These interactions allow the LLM to infer the customer's intent, based on their behavior within the session.
		
		\item \textit{Candidate Set}: A candidate set of products is supplied to the RS. The system then ranks these products, recommending the top-ranked items to the customer.
	\end{itemize}
	For clarity, we provide an example on the recommendation of the movie in Fig.~\ref{fig:rs_example} to illustrate how the proposed RS operates. In this example, the recommendation prompt shown is provided to the LLM, along with the session interactions and the candidate set. The LLM processes this information to generate a ranked list of items from the candidate set. The top-ranked items, specifically the top-5, are then recommended to the customer.
	
	\subsection{RSBench Optimization Problems}
	In this subsection, we introduce the details of RSBench. First, we define the decision variables and objective functions used in RSBench problems. Then, we present the nine multiobjective instances, which serve as a recommended set of benchmark problems for investigating the performance of LLM-based EAs in RSs.
	
	\subsubsection{Decision Variables}
	As discussed in Section~\ref{sec:task_description}, the LLM's recommendation is primarily guided by the recommendation prompt. Consequently, the quality and structure of the prompt have a significant impact on the performance of the RS. To maximize the system's performance, it is crucial to identify optimal recommendation prompts. Therefore, in the RSBench problem set, {the recommendation prompt, denoted as $\textbf{p}$, is treated as the decision variable}.
	
	\subsubsection{Objective Functions}
	Several indicators can be employed to evaluate the performance of the RS from various perspectives. In this paper, we primarily utilize the following three indicators to assess the RS, using them as objective functions.
	\begin{itemize}
		\item \textit{Recommendation Accuracy}: Recommendation accuracy is used to assess how effectively the RS identifies the customer's intentions and suggests the desired items~\cite{steck2011item}. An empirical estimate of recommendation accuracy can be calculated using the existing ground truth dataset. Given a dataset $\mathcal{D} = \{ ( \textbf{X}_{s}^{(l)}, \mathcal{Y}_{c}^{(l)}, \textbf{y}_{t}^{(l)} ) \}_{l=1}^{N}$, where $\textbf{X}_{s}^{(l)}$ is a sequence of session, $\mathcal{Y}_{c}^{(l)}$ is a set of candidate items, and $\textbf{y}_{t}^{(l)} \in \mathcal{Y}_{c}^{(l)}$ is a ground truth target item selected by the customer, we employ the following function to estimate recommendation accuracy
		\begin{equation}\label{eq:accuracy}
		\begin{aligned}
		F_{acc}(\textbf{p}) = \frac{1}{N} \cdot \sum_{l=1}^{N} \mathbb{I} \left[ R_{llm}(\textbf{y}_{t}^{(l)}|\textbf{X}_{s}^{(l)}, \mathcal{Y}_{c}^{(l)}, \textbf{p}) \leq K \right].
		\end{aligned}
		\end{equation}
		In \eqref{eq:accuracy}, $\mathbb{I}[\cdot]$ is an indicator function, and $R_{llm}(\textbf{y}_{t}^{(l)} | \textbf{X}_{s}^{(l)}, \mathcal{Y}_{c}^{(l)}, \textbf{p})$ is the predicted ranking of $\textbf{y}_{t}^{(l)}$ provided by the LLM with the input of $\textbf{X}_{s}^{(l)}$, $\mathcal{Y}_{c}^{(l)}$, and the recommendation prompt $\textbf{p}$. From \eqref{eq:accuracy}, we observe that recommendation accuracy considers a recommendation correct when the target item is among the top-\textit{K}-ranked items in the candidate set. The accuracy rate is then calculated as the indicator to evaluate the performance of the recommendation.
		
		\item \textit{Recommendation Diversity}: In many cases, the suggested items for customers may contain similar or overlapping information~\cite{zhang2008avoiding}. However, in real-world RSs, it is equally important to offer a diverse set of recommendations to keep users engaged by introducing novel content they may not have explicitly searched for but might still enjoy~\cite{kunaver2017diversity}. To achieve this, we aim to provide recommendations that exhibit a certain degree of diversity~\cite{yin2023understanding}. In this context, the information of an item is represented by its category, and the diversity is calculated as follows
		\begin{equation}\label{eq:diversity}
		\begin{aligned}
		F_{div}(\textbf{p}) = \frac{1}{N} \cdot \sum_{l=1}^{N} \frac{ | \cup_{k=1}^{K} C[ T^{k}_{llm} (\textbf{X}_{s}^{(l)}, \mathcal{Y}_{c}^{(l)}, \textbf{p}) ] | }{ \sum_{k=1}^{K} \left| C[ T^{k}_{llm} (\textbf{X}_{s}^{(l)}, \mathcal{Y}_{c}^{(l)}, \textbf{p}) ] \right| }
		\end{aligned}
		\end{equation}
		where $T^{k}_{llm} (\textbf{X}_{s}^{(l)}, \mathcal{Y}_{c}^{(l)})$ denotes the item ranked $k$ by the LLM-based RS, $C[\cdot]$ represents the category set associated with an item, and $|\cdot|$ denotes the size of a set.
		The formula first calculates the ratio of the number of unique categories among the top-$K$ items to the sum of the categories of each top-$K$ item for each recommendation sample. It then computes the average ratio across all \textit{N} recommendation samples. This formula reflects diversity by assigning smaller ratios to cases with high category overlap among top-$K$ items, which implies reduced diversity. Hence, higher values of $F_{div}$ indicate greater diversity in the recommendations.
		
		\item \textit{Recommendation Fairness}: Recommendation fairness is another important indicator considered in this paper~\cite{li2022fairness}. In RSs, biases can easily arise. For instance, the system may favor recommending popular items that have been available for a longer period, as these products typically exhibit higher click-through and purchase rates. Newly launched items, however, may lack sufficient historical data, causing them to be overlooked, even if they better meet customer needs~\cite{abdollahpouri2017controlling}. Therefore, ensuring fairness in recommendations is crucial. In this paper, we calculate recommendation fairness based on the click-through rates of the items~\cite{abdollahpouri2017controlling}. First, we divide all items into two sets: $\mathcal{H}$, which contains the top 20\% of items based on click-through rates, and $\mathcal{L}$, which includes the remaining 80\%. The recommendation fairness is then calculated as follows:
		\begin{equation}\label{eq:fariness}
		\begin{aligned}
		F_{fair}(\textbf{p}) = \frac{1}{N} \cdot \sum_{l=1}^{N} \frac{ \sum_{k=1}^{K} \mathbb{I}[ T_{llm}^k(\textbf{X}_{s}^{(l)}, \mathcal{Y}_{c}^{(l)}, \textbf{p}) \in \mathcal{L} ] }{ K }.
		\end{aligned}
		\end{equation}
		The formula calculates the ratio of items with lower click-through rates within the top-$K$ items for each recommendation sample, then computes the average ratio across $N$ samples. A higher $F_{fair}$ value indicates a greater tendency to recommend items with lower click-through rates, thus reflecting improved fairness in the recommendation outcomes.
	\end{itemize}
	
	\textbf{Remark}: It should be noted that all the objectives introduced above are maximization objectives.
	
	\subsubsection{Instances of RSBench}
	\begin{table}[!t]
		\caption{The Statistical Characteristics of Datasets. ``Density Indicator" Refers to the Average Frequency of Each Item Appearing in the Dataset.}\label{tab:datasets}
		\centering
		\resizebox{8.5cm}{!}{\begin{tabular}{c|cccc}
				\hline
				& Items & Sessions & Ave. Session Length & Density Indicator     \\
				\hline
				ML-1M     & 3416  & 784860 & 6.85 & 1573.86 \\
				Games     & 17389 & 100018 & 4.18 & 24.04   \\
				Bundle    & 14240 & 2376   & 6.73 & 1.12    \\
				\hline
		\end{tabular}}
	\end{table}
	
	Based on the definitions of the decision variables and objective functions provided above, we introduce nine instances of the RSBench problem set. To obtain the objective function values, datasets containing customer consumption and click data are required. In this benchmark, we use three real-world datasets from different domains. Specifically, MovieLens-1M (ML-1M)\footnote{https://grouplens.org/datasets/movielens/} is a movie dataset containing customer ratings for various films. The ``Games" dataset is a subcategory from the Amazon dataset~\cite{ni2019justifying}, which includes customer ratings for video games. The ``Bundle" dataset~\cite{sun2022revisiting} consists of session data from three Amazon subcategories: Electronics, Clothing, and Food. The statistical features of these datasets are summarized in Table~\ref{tab:datasets}. For the ML-1M and Games datasets, we chronologically order the rated items for each user and divide them into sessions by day. 
	
	\begin{table}[!t]
		\caption{The Settings of the RSBench Problems. ``Dataset" means the dataset used in the benchmark problem, ``Batch Size" represents the number of selected samples used for estimating the objective function, $m$ denotes the number of objectives, and ``Objective Functions" are the evaluation indicators considered in the benchmark problem.}\label{tab:benchmark}
		\centering
		\resizebox{8.5cm}{!}{\begin{tabular}{c|c|c|c|c}
				\hline
				& Dataset & Batch Size & $m$ & Objective Functions \\
				\hline
				RSBench-1 & ML-1M  & 10 & 2 & $F_{acc}, F_{div}$ \\
				RSBench-2 & Games  & 10 & 2 & $F_{acc}, F_{div}$ \\
				RSBench-3 & Bundle & 10 & 2 & $F_{acc}, F_{div}$ \\
				RSBench-4 & ML-1M  & 10 & 2 & $F_{acc}, F_{fair}$ \\
				RSBench-5 & Games  & 10 & 2 & $F_{acc}, F_{fair}$ \\
				RSBench-6 & Bundle & 10 & 2 & $F_{acc}, F_{fair}$ \\
				RSBench-7 & ML-1M  & 10 & 3 & $F_{acc}, F_{div}, F_{fair}$ \\
				RSBench-8 & Games  & 10 & 3 & $F_{acc}, F_{div}, F_{fair}$ \\
				RSBench-9 & Bundle & 10 & 3 & $F_{acc}, F_{div}, F_{fair}$ \\
				\hline
		\end{tabular}}
	\end{table}
	
	Using these three datasets, the settings for the nine RSBench instances are presented in Table~\ref{tab:benchmark}. All RSBench problems are multiobjective optimization problems. Note that the LLM-based RS requires continuous online interaction with the LLM during the recommendation process. As this interaction process is time-consuming, it is impractical to evaluate the objectives using the entire training set. Therefore, when calculating the objective function values, we randomly select a batch of samples from the training set and compute the results based on this batch. As shown in Table~\ref{tab:benchmark}, the batch size for all RSBench problems in this paper is set to 10. However, users of this benchmark may choose a larger batch size for more accurate objective function estimation, although this may result in longer evaluation times. Developing a more efficient and time-saving evaluation strategy is a promising direction for further exploration, though it is beyond the scope of this paper.
	
	{\color{blue}
		\subsubsection{Challenges of the RSBench Problems}
		In general, we consider that the RSBench problems mainly contains the following three challenges:
		\begin{itemize}
			\item {\textit{High Computational Costs}}: As shown in \eqref{eq:accuracy}, \eqref{eq:diversity}, and \eqref{eq:fariness}, all objective values are evaluated using the Monte Carlo method. Consequently, computing the objective function requires testing each sample individually and repeatly. This frequent interaction with APIs incurs significant computational costs in terms of time and token consumption, making RSBench problems inherently expensive. When applying EAs to RSBench, this computational expense must be carefully considered to ensure efficient optimization. 
			\item \textit{Uncertainty}: Additionally, the evaluation of objective functions introduces uncertainty in assessing the true performance of recommendation prompts. This uncertainty arises from three primary sources. First, the Monte Carlo evaluation may deviate from the actual performance of a prompt, introducing potential bias. Second, the stochastic nature of LLM inference leads to variability across repeated trials. Third, users themselves may not always exhibit clear or consistent preferences, further embedding uncertainty into RSBench. Such uncertainties can misguide the evolutionary process, ultimately hindering convergence and degrading overall performance.
			\item \textit{Searching in the Language Space}: In the RSBench problem, our objective is to identify an optimal recommendation prompt to guide the LLM in completing the recommendation task. Unlike conventional EAs, which search for optimal solutions in a fixed-dimensional vector space, RSBench requires searching for optimal solutions in a language space. This unique search space introduces two key challenges. First, since the solutions are expressed in natural language, they must maintain linguistic coherence, which imposes implicit constraints on word selection and arrangement. Second, as natural language does not have a fixed structure or length, the optimization process must operate within a variable-length search space, requiring adaptive strategies to effectively explore and refine solutions.
		\end{itemize}
	}
	
	\section{Language Model Multiobjective Evolutionary Algorithms for RSBench}
	In this section, we present the details of the LLM-based EAs employed for RSBench in this paper. First, we describe two fundamental operations utilized in the algorithms: the initialization of the recommendation prompt population and the generation of offspring recommendation prompts. Next, we provide a detailed explanation of the three LLM-based EAs.
	
	\begin{figure}[!t]
		\begin{center}
			\includegraphics[width=1\columnwidth]{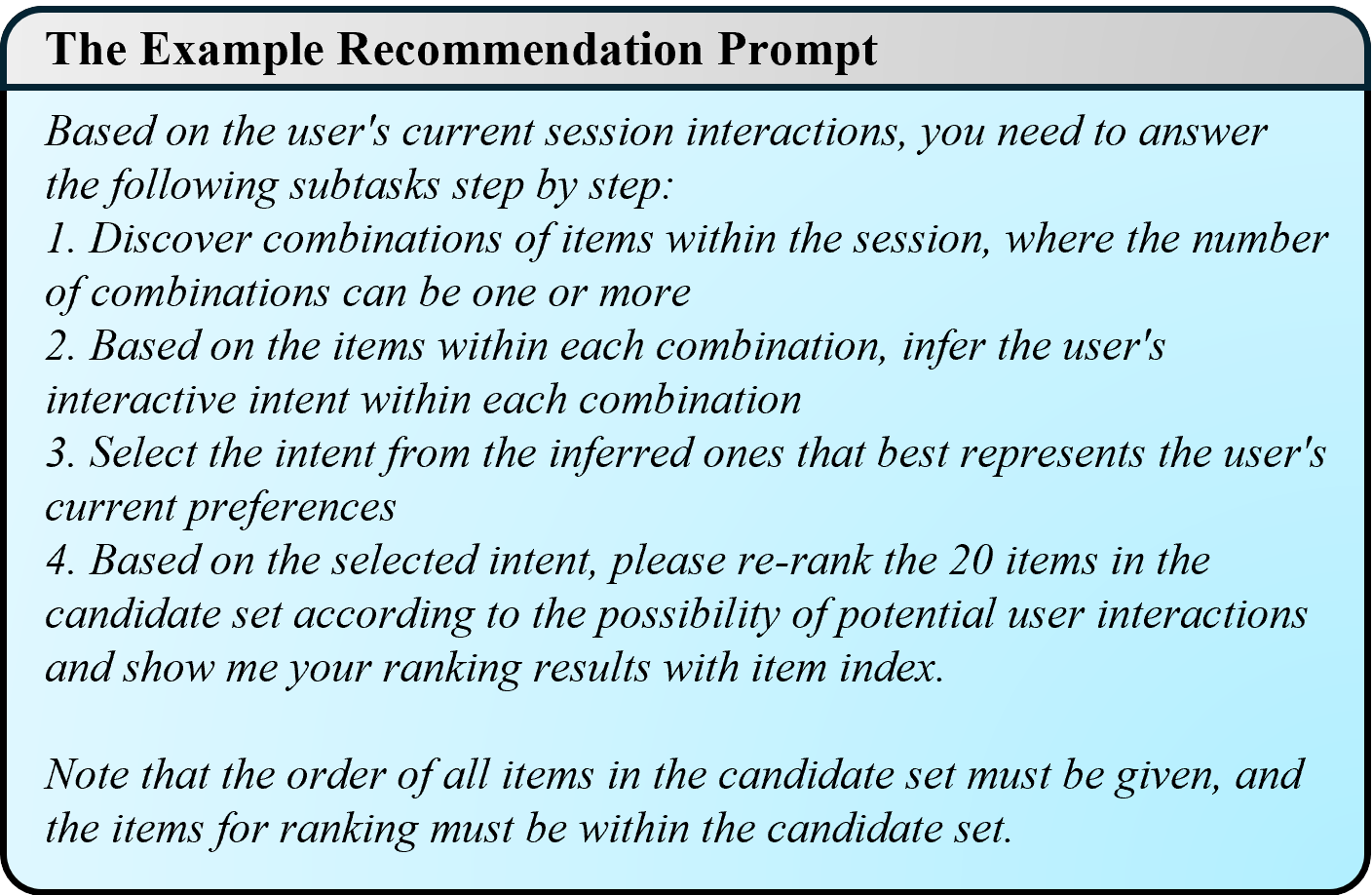}
			\vspace{-5mm}
			\caption{The example recommendation prompt designed following the chain-of-thought.}
			\label{fig:example_prompt}
		\end{center}
	\end{figure}
	
	\begin{figure}[!t]
		\begin{center}
			\includegraphics[width=1\columnwidth]{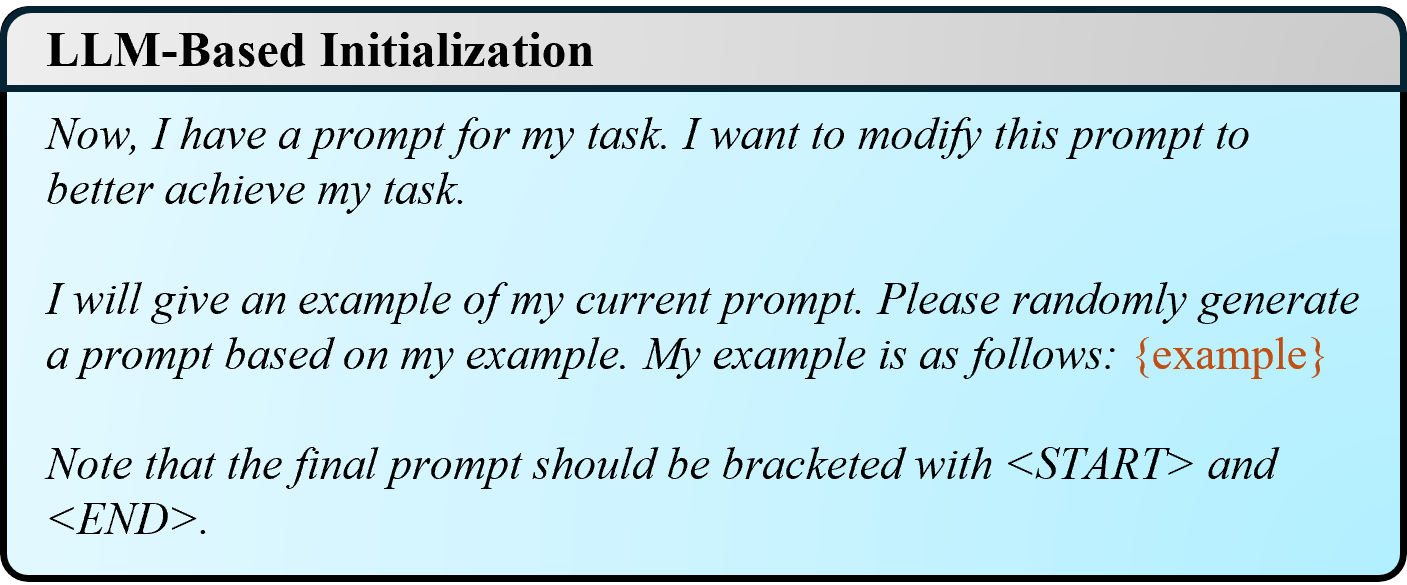}
			\vspace{-5mm}
			\caption{The prompt for the initialization of the population.}
			\label{fig:llm_initialization}
		\end{center}
	\end{figure}
	
	\subsection{Basic Operations}\label{sec:operations}
	\subsubsection{Initialization of the Recommendation Prompt Population}\label{sec:initialization}
	
	While the robust capabilities of LLMs enable them to excel at various downstream tasks, their performance is highly dependent on the quality of the prompts provided. Over the past year, numerous prompt structures, such as chain-of-thought~\cite{wei2022chain} and tree-of-thought~\cite{yao2024tree}, have been proposed to improve LLM performance. In this paper, to expedite the discovery of high-quality prompts, we introduce a well-designed recommendation prompt as the example based on the chain-of-thought approach. The details of this example recommendation prompt are presented in Fig.~\ref{fig:example_prompt}. 
	{
		This example recommendation prompt guides the recommendation process through four structured steps. First, it identifies meaningful item combinations within the session to capture contextual relationships. Second, it infers the user's interactive intent based on these combinations. Third, it selects the most relevant intent that best represents the user's preferences. Finally, it re-ranks the candidate items accordingly. By systematically breaking down user interactions, this prompt ensures that recommendations are context-aware and preference-driven.
	}
	The initial population of recommendation prompts is subsequently generated using the LLM based on the example recommoned prompt, with the initialization prompt shown in Fig.~\ref{fig:llm_initialization}. In this manner, $N$ recommendation prompts are produced to form the initial population. {Such an initialization process ensures, on the one hand, that the recommendation prompts in the generated initial population are logically structured to guide the LLM in effectively completing the recommendation task. On the other hand, it also introduces a certain degree of diversity within the population.}
	
	\subsubsection{Offspring Recommendation Prompts Generation}\label{sec:crossover}
	\begin{figure}[!t]
		\begin{center}
			\includegraphics[width=1\columnwidth]{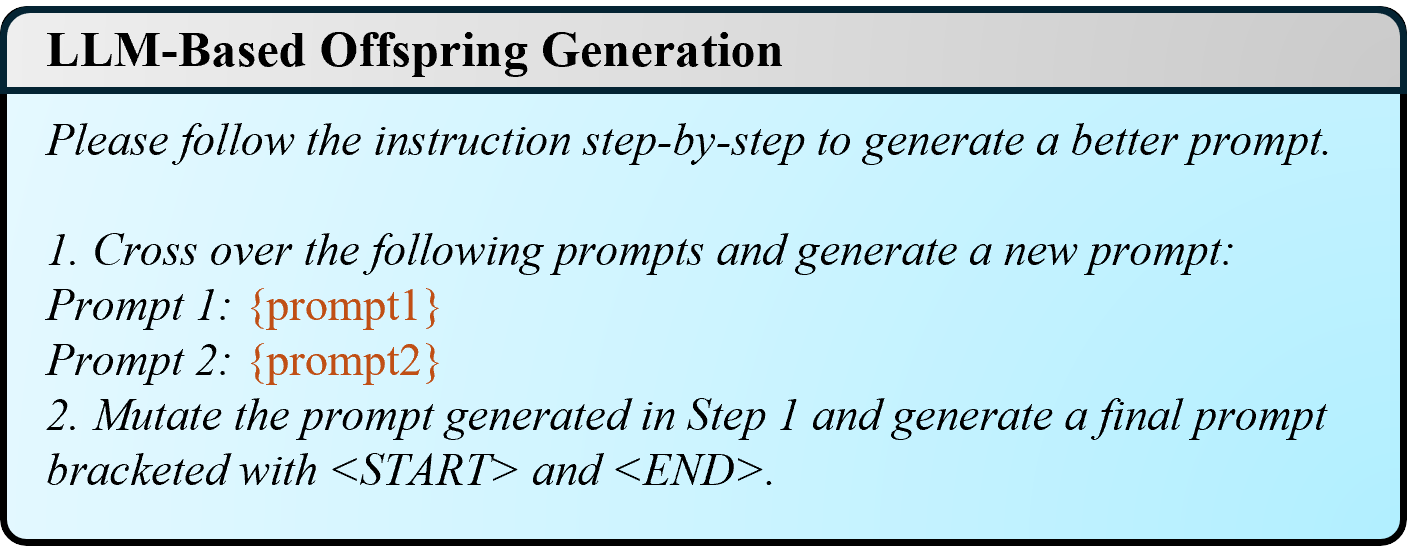}
			\vspace{-5mm}
			\caption{The prompt for the offspring generation.}
			\label{fig:crossover}
		\end{center}
	\end{figure}
	
	In traditional EAs, crossover and mutation operators are used to generate offspring solutions. However, these operators are typically designed for continuous or discrete vectors and are not well-suited for handling natural language, such as recommendation prompts. Recent studies have demonstrated that, with appropriate prompts, LLMs can themselves serve as effective crossover and mutation operators for natural language processing tasks~\cite{guo2023connecting}. Following this approach, we employ an LLM-based crossover and mutation mechanism in this work. The prompt used for the LLM-based crossover and mutation operator is illustrated in {Fig.~\ref{fig:crossover}}. {This prompt simulates the crossover and mutation process in conventional EAs. Specifically, the LLM performs crossover by combining two parent recommendation prompts, ensuring a diverse set of base recommendation prompts. Subsequently, LLM-based mutation employs conditional generation to fine-tune the newly generated prompts, further enhancing their diversity. Compared to directly fine-tuning the recommendation prompts, this approach tends to achieve better population diversity. Consequently, the LLM-based operator facilitates the effective generation of an offspring population from the parent population.}
	
	\subsection{Main Frameworks}
	By incorporating the operations introduced in Section~\ref{sec:operations} into multiobjective evolutionary frameworks, LLM-driven multiobjective EAs can be developed. In this paper, we consider three representative EA frameworks: NSGA-II, MOEA/D, and IBEA. By integrating LLM-based operations, we derive three LLM-based EAs, denoted as LLM-NSGA-II, LLM-MOEA/D, and LLM-IBEA, respectively.
	
	\subsubsection{LLM-NSGA-II}
	\begin{algorithm}[t]
		
		\caption{\textit{LLM-NSGA-II}}\label{Alg:LLM-NSGA2}
		\renewcommand{\algorithmicrequire}{\textbf{Input:}} 
		\renewcommand{\algorithmicensure}{\textbf{Output:}}
		\begin{algorithmic}[1]
			\REQUIRE The population size $N$; objective functions $\textbf{F}(\textbf{p})$; 
			\ENSURE The nondominated recommendation prompt set $\mathcal{P}$;
			\STATE Generate the initial population $\mathcal{P} = \{ \textbf{p}_1, \ldots, \textbf{p}_N \}$ by using the initial operations introduced in Section~\ref{sec:initialization}, and evaluate $\mathcal{P}$ by using the objective functions;
			\STATE $t \leftarrow 0$
			\WHILE{$t < t_{max}$}
			\STATE Generate an offspring population of recommendation prompts $\mathcal{Q} = \{ \textbf{q}_1, \ldots, \textbf{q}_N \}$ according to the operations introduced in Section~\ref{sec:crossover};
			\STATE Utilizing the environmental selection of NSGA-II to select recommendation prompts from $\mathcal{P} \cup \mathcal{Q}$, thus forming the new population $\mathcal{P}$;
			\STATE $t \leftarrow t + 1$;
			\ENDWHILE
			\RETURN A population of nondominated recommendation prompts $\mathcal{P}$.
		\end{algorithmic}
	\end{algorithm}
	
	The LLM-NSGA-II follows the classical NSGA-II framework, with its pseudo-code presented in \textbf{Algorithm}~\ref{Alg:LLM-NSGA2}. Initially, a population $\mathcal{P}$ of $N$ recommendation prompts is generated using the initialization operation described in Section\ref{sec:initialization}, and their performance is evaluated using the defined objective functions. In each iteration, an offspring population $\mathcal{Q}$ is generated using the crossover operation introduced in Section~\ref{sec:crossover}. The offspring prompts are then evaluated based on the same objective functions, and elitist selection is performed to choose $N$ recommendation prompts from the combined population $\mathcal{P} \cup \mathcal{Q}$ using nondominated sorting and crowding distance~\cite{nsga2}. These selected prompts form the new population for the next iteration. Upon satisfying the termination conditions, the final population of nondominated recommendation prompts is returned.
	
	\subsubsection{LLM-MOEA/D}
	\begin{algorithm}[t]
		
		\caption{\textit{LLM-MOEA/D}}\label{Alg:MOEAD}
		\renewcommand{\algorithmicrequire}{\textbf{Input:}} 
		\renewcommand{\algorithmicensure}{\textbf{Output:}}
		\begin{algorithmic}[1]
			\REQUIRE The population size $N$; objective functions $\textbf{F}(\textbf{p})$; set of decomposition weight vectors $\mathcal{W} = \{ \textbf{w}_1,\ldots,\textbf{w}_N \}$; 
			\ENSURE The nondominated recommendation prompt set $\mathcal{P}$;
			\STATE Generate the initial population $\mathcal{P} = \{ \textbf{p}_1, \ldots, \textbf{p}_N \}$ by using the initial operations introduced in Section~\ref{sec:initialization}, and evaluate $\mathcal{P}$ by using the objective functions;
			\STATE Generate the neighborhood set $\mathcal{B}_{i}$, the external population $\mathcal{E}$, and the reference point $\textbf{z} = (z_1, . . . , z_m)^T$ following the MOEA/D framework;
			\STATE $t \leftarrow 0$
			\WHILE{$t < t_{max}$}
			\FOR{$i = \{ 1, \ldots, N \}$} 
			\STATE Generate an offspring $\textbf{p}^{new}$ following the MOEA/D framework by using the LLM-based offspring generation introduced in Section~\ref{sec:crossover}, and evaluate $\textbf{p}^{new}$ by using the objective functions;
			\STATE Update the reference point $\textbf{z}$, the population $\mathcal{P}$ and the external population $\mathcal{E}$ following the MOEA/D framework;
			\ENDFOR
			\STATE $t \leftarrow t + 1$;
			\ENDWHILE
			\RETURN An external population of nondominated recommendation prompts $\mathcal{E}$.
		\end{algorithmic}
	\end{algorithm}
	
	Decomposition-based evolutionary multiobjective optimization is a classical paradigm in multiobjective optimization. LLM-MOEA/D follows the decomposition-based method, with the pseudo-code presented in \textbf{Algorithm}~\ref{Alg:MOEAD}. As in the MOEA/D framework, a neighborhood set $\mathcal{B}_{i}$ is initialized for each weight vector $\textbf{w}_i$. The initial population $\mathcal{P}$ is generated using the LLM-based initialization described in Section~\ref{sec:initialization}, and all recommendation prompts in $\mathcal{P}$ are evaluated using the objective functions. Based on these evaluations, the external population $\mathcal{E}$ and the reference point $\textbf{z}$ are initialized. During the iteration process, for each weight vector, an offspring recommendation prompt $\textbf{p}^{new}$ is produced using the LLM-based offspring generation method detailed in Section~\ref{sec:crossover}. After evaluating $\textbf{p}^{new}$, the reference point $\textbf{z}$, the population $\mathcal{P}$, and the external population $\mathcal{E}$ are updated in accordance with the MOEA/D framework~\cite{moead}. Once the termination conditions are satisfied, the external population $\mathcal{E}$ of nondominated recommendation prompts is returned.
	
	\subsubsection{LLM-IBEA}
	\begin{algorithm}[t]
		
		\caption{\textit{LLM-IBEA}}\label{Alg:IBEA}
		\renewcommand{\algorithmicrequire}{\textbf{Input:}} 
		\renewcommand{\algorithmicensure}{\textbf{Output:}}
		\begin{algorithmic}[1]
			\REQUIRE The population size $N$; objective functions $\textbf{F}(\textbf{p})$; 
			\ENSURE The nondominated recommendation prompt set $\mathcal{P}$;
			\STATE Generate the initial population $\mathcal{P} = \{ \textbf{p}_1, \ldots, \textbf{p}_N \}$ by using the initial operations introduced in Section~\ref{sec:initialization}, and evaluate $\mathcal{P}$ by using the objective functions;
			\STATE $t \leftarrow 0$
			\WHILE{$t < t_{max}$}
			\STATE Generate an offspring population of recommendation prompts $\mathcal{Q} = \{ \textbf{q}_1, \ldots, \textbf{q}_N \}$ according to the operations introduced in Section~\ref{sec:crossover};
			\STATE Utilizing the environmental selection of IBEA to select recommendation prompts from $\mathcal{P} \cup \mathcal{Q}$, thus forming the new population $\mathcal{P}$;
			\STATE $t \leftarrow t + 1$;
			\ENDWHILE
			\RETURN A population of nondominated recommendation prompts $\mathcal{P}$.
		\end{algorithmic}
	\end{algorithm}
	
	The most classical indicator-based evolutionary algorithm, i.e., IBEA~\cite{ibea}, is used to drive the evolution of LLM-IBEA. Similar to the previous two algorithms, it begins by initializing a population $\mathcal{P}$ using the LLM-based initialization process. The offspring population $\mathcal{Q}$ is then generated through the LLM-based crossover and mutation. After evaluating the offspring recommendation prompts, the $I_{\epsilon^{(+)}}$ indicator-based environmental selection~\cite{ibea} is applied to select $N$ recommendation prompts, forming a new population for the next iteration. Once the termination conditions are met, the final population of nondominated recommendation prompts is returned. The overall process of the LLM-IBEA is shown in \textbf{Algorithm}~\ref{Alg:IBEA}
	
	\begin{table*}[!h]
		\centering
		\caption{HV results of LLM-NSGA-II, LLM-MOEA/D, and LLM-IBEA on RSBench Problems, averaged over 5 Independent Runs with Diverse Generated Candidate Sets.}\label{tab:results}
		\begin{tabular}{c|c|c|c||c|c|cccc}
			\hline
			\multirow{3}{*}{Problems} & \multicolumn{3}{c||}{Training Sets} & \multicolumn{3}{c}{Validation Sets} \\
			\cline{2-7}
			& {LLM-NSGA-II} & {LLM-MOEA/D} & {LLM-IBEA} & {LLM-NSGA-II} & {LLM-MOEA/D} & {LLM-IBEA} \\
			\cline{2-7}
			& \multicolumn{1}{c|}{Average HV $\pm$ Std}            & \multicolumn{1}{c|}{Average HV $\pm$ Std}         & \multicolumn{1}{c||}{Average HV $\pm$ Std}  & \multicolumn{1}{c|}{Average HV $\pm$ Std}                & \multicolumn{1}{c|}{Average HV $\pm$ Std}         & \multicolumn{1}{c}{Average HV $\pm$ Std}  \\
			\hline
			RSBench-1 & 0.6606$\pm$0.0145          & 0.6486$\pm$0.0242          & \textbf{0.7088}$\pm$\textbf{0.0045} & 0.4542$\pm$0.0171          & 0.4450$\pm$0.0299          & \textbf{0.5060}$\pm$\textbf{0.0160} \\
			RSBench-2 & 0.5901$\pm$0.0333          & 0.6354$\pm$0.0641          & \textbf{0.6694}$\pm$\textbf{0.0386} & 0.4153$\pm$0.0111          & 0.4081$\pm$0.0212          & \textbf{0.4293}$\pm$\textbf{0.0430} \\
			RSBench-3 & 0.7637$\pm$0.0691          & \textbf{0.7650}$\pm$\textbf{0.0155} & 0.7448$\pm$0.0235          & \textbf{0.5011}$\pm$ \textbf{0.0095} & 0.4925$\pm$0.0301          & 0.4621$\pm$0.0075          \\
			RSBench-4 & 0.4169$\pm$0.0284          & \textbf{0.5420}$\pm$\textbf{0.0175} & 0.5137$\pm$0.0413          & 0.2920$\pm$0.0326          & \textbf{0.3227}$\pm$\textbf{0.0193} & 0.3061$\pm$0.0149          \\
			RSBench-5 & 0.4143$\pm$0.0263          & 0.3470$\pm$0.0333          & \textbf{0.4171}$\pm$\textbf{0.0220} & 0.2090$\pm$0.0073          & 0.2019$\pm$0.0119          & \textbf{0.2146}$\pm$\textbf{0.0130} \\
			RSBench-6 & 0.2680$\pm$0.0273          & 0.2656$\pm$0.0366          & \textbf{0.2842}$\pm$\textbf{0.0231} & 0.1291$\pm$0.0011          & 0.1158$\pm$0.0011          & \textbf{0.1306}$\pm$\textbf{0.0054} \\
			RSBench-7 & 0.3511$\pm$0.0113          & 0.3869$\pm$0.0119          & \textbf{0.4116}$\pm$\textbf{0.0281} & 0.2442$\pm$0.0059          & \textbf{0.2500}$\pm$\textbf{0.0265} & 0.2412$\pm$0.0206          \\
			RSBench-8 & \textbf{0.2834}$\pm$\textbf{0.0238} & 0.2647$\pm$0.0144          & 0.2676$\pm$0.0250          & \textbf{0.1444}$\pm$ \textbf{0.0164} & 0.1272$\pm$0.0052          & 0.1438$\pm$0.0037          \\
			RSBench-9 & 0.2580$\pm$0.0291          & 0.2641$\pm$0.0437          & \textbf{0.2652}$\pm$\textbf{0.0318} & 0.1133$\pm$0.0048          & \textbf{0.1210}$\pm$\textbf{0.0009} & 0.1185$\pm$0.0036 \\
			\hline
		\end{tabular}
	\end{table*}
	
	\begin{figure*}[!t]
		\begin{center}
			\subfigure[]{\includegraphics[width=0.52\columnwidth]{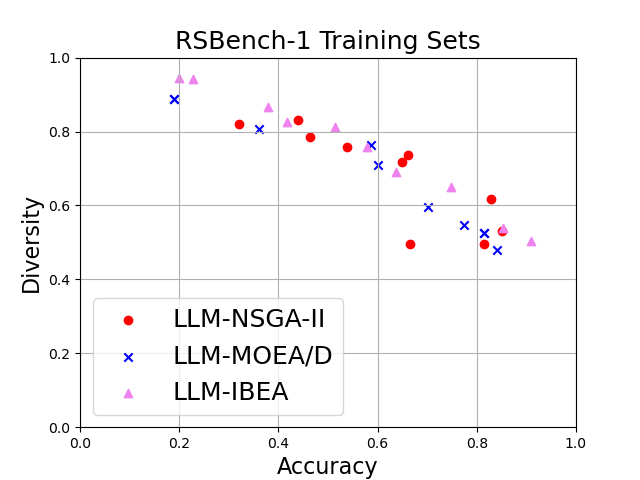}}
			\subfigure[]{\includegraphics[width=0.52\columnwidth]{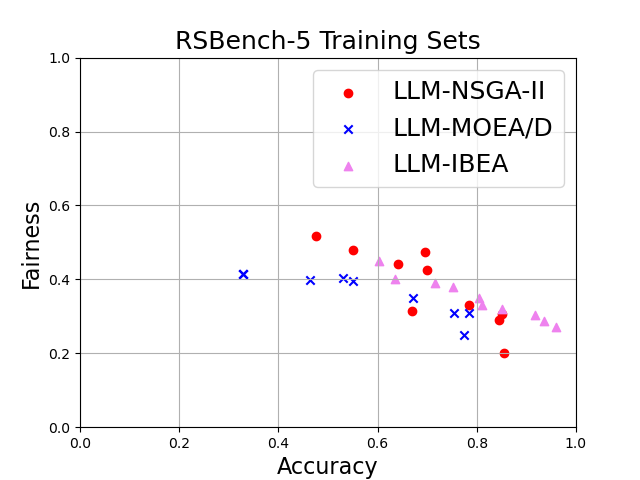}}
			\subfigure[]{\includegraphics[width=0.52\columnwidth]{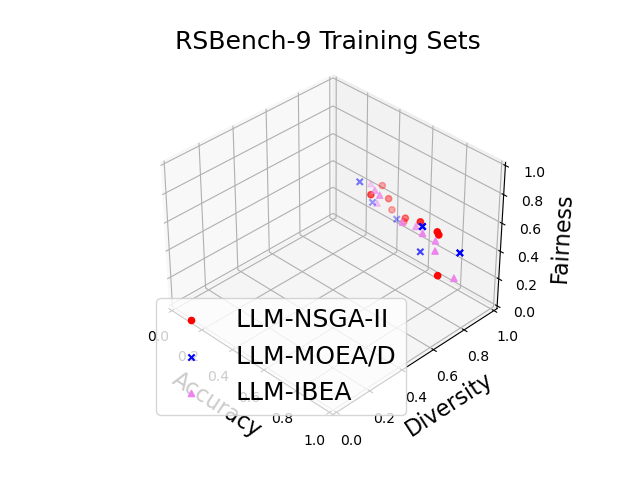}}
			\caption{Pareto front approximation results provided by LLM-NSGA-II, LLM-MOEA/D, and LLM-IBEA on RSBench-1, RSBench-5, and RSBench-9 with the randomly sampled batch in training set. (a) Results on RSBench-1. (b) Results on RSBench-5. (c) Results on RSBench-9.}
			\label{fig:training_set_results}
		\end{center}
	\end{figure*}
	
	\section{Experimental Studies}
	
	\subsection{Parameter Settings and Implementation Details}
	In our experiment studies, GPT-3.5 {(version \verb|gpt-3.5-turbo|)}~\cite{OpenAI} is employed in both the RSBench and the LLM-based EAs, {and the temperature of the LLM is set to 0.7}. For the algorithms LLM-NSGA-II, LLM-MOEA/D, and LLM-IBEA, we set the maximum number of iterations to 20 and the population size to 10. In LLM-MOEA/D, 10 weight vectors are generated using the Riesz s-energy method~\cite{ref_dirs_energy}, with a {neighborhood size of 3. (The influence of the neighborhood size is investigated in Section V-G.)} The environmental selection mechanisms for LLM-NSGA-II, LLM-MOEA/D, and LLM-IBEA adhere to those specified in the original papers~\cite{nsga2, moead, ibea}. 
	{To evaluate the performance of the algorithms, we partitioned each dataset} {into training and validation sets. Specifically, we randomly split the data in an 8:2 ratio, with 80\% allocated to the training set and the remaining 20\% to the validation set.}
	In each iteration, a batch of 10 samples is randomly selected from the training set to evaluate objective function values. For each sample, 20 items are saved in the candidate set; if the target item ranks within the top 10 after re-ranking by the RS, the recommendation is deemed correct. The optimization process is repeated five times, each with a different random seed (i.e., 0, 10, 42, 625, and 2023) to generate diverse candidate sets. {For the validation set evaluation, we randomly select 200} samples to estimate the objective functions, maintaining all other settings consistent with those used for the training set. {Both the RSBench and the LLM-based EAs were implemented in a Windows 10 Professional environment, using an Intel(R) Xeon(R) CPU ES-1650 with a clock speed of 3.6 GHz and 64 GB of RAM. The implementation was carried out using Python version 3.10.2.}

	\subsection{Hypervolume Measure}
	We evaluated the algorithm quality using the {hypervolume (HV) metric~\cite{auger2012hypervolume}}, which measures the size of the objective space region dominated by a set of solutions. Since the objective values in the RSBench problems are maximization problems and confined to the region $[0,1]$, we use the point $(0,0)$ as the reference for two-objective problems (i.e., RSBench-1 to RSBench-6) and $(0,0,0)$ as the reference for three-objective problems (i.e., RSBench-7 to RSBench-9).
	
	\subsection{Results and Insights of LLM-Based EAs on RSBench}
	\begin{figure*}[!t]
		\begin{center}
			\subfigure[]{\includegraphics[width=0.52\columnwidth]{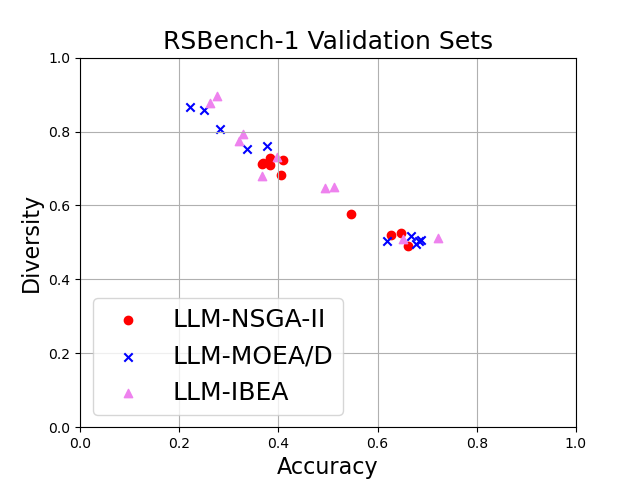}}
			\subfigure[]{\includegraphics[width=0.52\columnwidth]{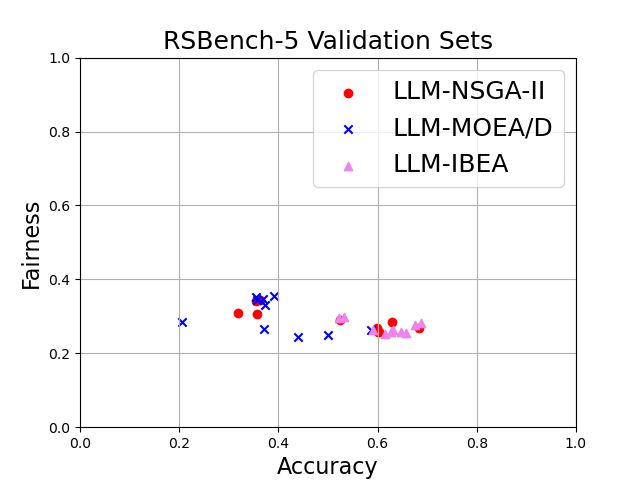}}
			\subfigure[]{\includegraphics[width=0.52\columnwidth]{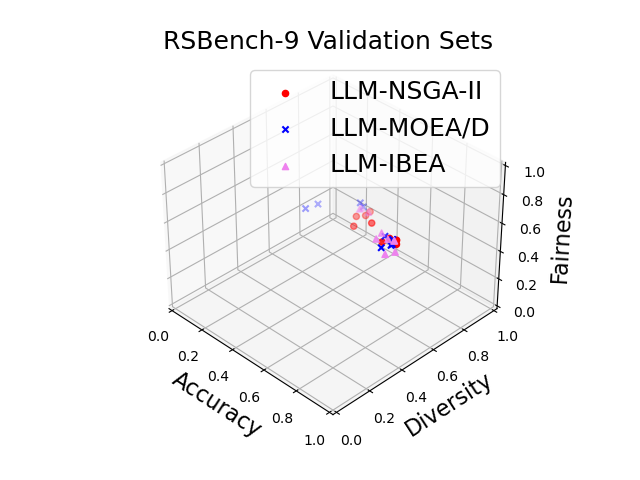}}
			\caption{Pareto front approximation results provided by LLM-NSGA-II, LLM-MOEA/D, and LLM-IBEA on RSBench-1, RSBench-5, and RSBench-9 with the randomly sampled batch in validation set. (a) Results on RSBench-1. (b) Results on RSBench-5. (c) Results on RSBench-9.}
			\label{fig:validation_set_results}
		\end{center}
	\end{figure*}
	
	In this subsection, we evaluate the performance of LLM-NSGA-II, LLM-MOEA/D, and LLM-IBEA on RSBench problems. Table~\ref{tab:results} presents the HV results obtained by these algorithms across multiple candidate sets generated with different random seeds. {The results indicate that LLM-IBEA achieves the best overall performance, attaining the highest HV values on six RSBench problems (RSBench-1, RSBench-2, RSBench-5 to RSBench-7, and RSBench-9) with the training sets and on four problems (RSBench-1, RSBench-2, RSBench-5, and RSBench-6) with the validation sets. LLM-NSGA-II and LLM-MOEA/D exhibit comparable performance. LLM-NSGA-II attains the highest HV value on one RSBench problem in the training set and two in the validation set, specifically excelling on RSBench-8 (training set) and RSBench-3 and RSBench-8 (validation set). LLM-MOEA/D achieves the highest HV values on two RSBench problems in the training set (RSBench-3 and RSBench-4) and three in the validation set (RSBench-4, RSBench-7, and RSBench-9).}
	
	Fig.~\ref{fig:training_set_results} and~\ref{fig:validation_set_results} illustrate the objective-space approximate points for nondominated recommendation prompts generated by LLM-NSGA-II, LLM-MOEA/D, and LLM-IBEA on RSBench-1, RSBench-5, and RSBench-9 for both the training and validation sets. As shown, all three algorithms produce recommendation prompts that are well distributed across the objective space and display complementary dominance characteristics. This distribution enables users to choose high-quality recommendation prompts that best align with their personal preferences, thereby enhancing customization.

	{\color{blue}In addition to the above experimental results, several key insights have been gained to better inform algorithm design in this area:
	\begin{itemize}		
		\item {Decomposition-based methods exhibit relatively unstable performance compared to the other two approaches. A possible explanation is that crossover and mutation operations are conducted within subpopulations that lack sufficient diversity. As a result, the LLM encounters difficulties in generating novel prompts through the recombination of candidate prompts within these limited diverse subpopulations, which restricts the algorithm’s exploration capability. This reduced exploratory ability may cause the prompts to partially converge to suboptimal solutions in certain runs, leading to instability in the optimization outcomes.} To examine this, we measure the diversity of prompts in the final population using the BERTScore. Specifically, we compute the BERTScore for each pair of recommender prompts in the final population and use the average score as a diversity metric. The results indicate that LLM-NSGA-II and LLM-IBEA achieve average scores of 0.1005 and 0.1025, respectively, whereas LLM-MOEA/D yields a larger score of 0.1036, reflecting reduced population diversity.
	
		\item Moreover, from the HV results obtained by the three algorithms, we found that the benchmark based on the Bundle dataset proves more difficult to optimize. This may be because RSs relying solely on prompts and LLMs in a zero-shot setting struggle to handle complex data involving diverse items within a bundle.
		
		\item Extending from two-objective to three-objective optimization problems poses additional challenges for all algorithms. This difficulty likely arises from the increased complexity of balancing trade-offs among accuracy, diversity, and fairness simultaneously. This may be obversed from Fig. 6 and Fig. 7. The obtained solutions exhibit an approximately degenerate Pareto front in the target space, making it difficult to generate recommendation prompts that explore regions with low fairness and diversity. This phenomenon is also a characteristic of RSBench with three objectives. In general, if a recommendation result achieves high fairness, less popular items are more likely to appear in the recommendation list. When these rare items are recommended alongside popular ones, the results typically also exhibit high diversity. This positive correlation arises because fairness promotes exposure of underrepresented items, while diversity is naturally enhanced when both frequent and infrequent items are included in the same recommendation list.
		
	\end{itemize}

	Moreover, from a recommendation perspective, a unique advantage of multiobjective evolutionary algorithms lies in their ability to generate a diverse set of prompts that capture trade-offs across different objectives. This diversity enables users to flexibly select the prompts that best align with their preferences, offering greater adaptability compared to non-EA-based prompt optimization strategies, which typically provide fewer options. We also compared two representative prompts obtained from LLM-IBEA to examine their impact on recommendation outcomes. We first give the two representative prompts:
	\begin{itemize}
		\item \textit{ 
			\textbf{Prompt A (higher diversity, lower accuracy):} Identifying unique combinations of items present in the session is crucial. Analyze the patterns of sequential user actions within the session and identify any recurring sequences. We analyze the items within each combination to deduce the user's specific interaction intent for that combination. Next, we determine the most appropriate user intent from the inferred intents that align with the user's current preferences. Predict the user's next intended action based on the patterns observed in the sequential data. Utilizing the selected intent, we reevaluate the 20 items in the candidate set for potential user interactions and present the revised ranking results including the item indices while maintaining the order of all items in the candidate set exclusively. Determine the most prominent user intent based on the analysis of patterns and interactions. Determine the likelihood of successful completion for each predicted action. Reorder the 20 items in the candidate set according to the likelihood of alignment with the identified intent, providing the revised ranking results with item indices. Prioritize the predicted actions based on the likelihood of successful completion and present the top 3 recommendations to the user.
		}
	
		\item \textit{	  
			\textbf{Prompt B (higher accuracy, lower diversity):} Identifying patterns in the user's session interactions to reveal any recurring themes or preferences is crucial. We need to analyze the items within each combination to deduce the user's specific interaction intent for that combination. Determine the most relevant intent from the inferred ones that align with the user's current preferences. Next, we analyze the relationship between the identified patterns and the items accessed to understand the user's intent behind each interaction. Utilizing the selected intent, we reorder the 20 items in the candidate set based on the likelihood of alignment with the identified intent, providing the revised ranking results with item indices. Ensure that the order of all items in the candidate set is maintained, and the items for ranking are exclusively from the candidate set.
		}
	\end{itemize}
	\textit{Prompt A} emphasizes repeated inference of user intent and multiple re-ranking operations within the candidate set. This design introduces broader coverage of potential user preferences and consequently increases the diversity of the recommendations. However, the repeated reasoning and ranking steps also add instability and noise, which can reduce accuracy. In contrast, \textit{Prompt B} adopts a more streamlined structure. It focuses on identifying a single most relevant intent and directly reorders the candidate set based on the likelihood of alignment with this intent. By limiting the process to one intent extraction and strictly constraining the output to the candidate set, \textit{Prompt B} yields more accurate recommendations. Nevertheless, this design reduces the exploratory capacity of the model, resulting in lower diversity in the final outcomes.
	}

	\subsection{Time and Token Consumptions on RSBench Problems}
	
	\begin{table}[!t]
		\centering
		\caption{Consumed time of LLM-NSGA-II, LLM-MOEA/D, and LLM-IBEA on RSBench Problems, averaged over 5 independent runs with diverse generated candidate sets.}\label{tab:time}
		\resizebox{8cm}{!}{\begin{tabular}{c|c|c|c}
				\hline
				\multirow{2}{*}{Problems} & LLM-NSGA-II & LLM-MOEA/D & LLM-IBEA \\
				\cline{2-4}
				& \multicolumn{3}{c}{Average Consumed Time (seconds)}              \\
				\hline
				RSBench-1 & 11565.23    & 9583.45    & 10536.54 \\
				RSBench-2 & 14879.32    & 13439.38   & 14358.89 \\
				RSBench-3 & 28875.37    & 25673.86   & 21603.25 \\
				RSBench-4 & 12669.37    & 9498.68    & 10811.16 \\
				RSBench-5 & 13543.74    & 10783.86   & 13733.42 \\
				RSBench-6 & 24326.51    & 20826.94   & 24127.54 \\
				RSBench-7 & 11720.80    & 10425.90   & 14535.28 \\
				RSBench-8 & 17261.30    & 13521.75   & 14299.40 \\
				RSBench-9 & 16191.24    & 19020.70   & 20244.55 \\
				\hline
		\end{tabular}}
	\end{table}
	
	\begin{table}[!t]
		\centering
		\caption{Consumed tokens of LLM-NSGA-II, LLM-MOEA/D, and LLM-IBEA on RSBench Problems, averaged over 5 Independent Runs with Diverse Generated Candidate Sets.}\label{tab:token}
		\begin{tabular}{c|c|c|c}
			\hline
			\multirow{2}{*}{Problems}  & LLM-NSGA-II & LLM-MOEA/D & LLM-IBEA \\
			\cline{2-4}
			& \multicolumn{3}{c}{Average Consumed Tokens} \\
			\hline
			RSBench-1 & 2831028 & 2249745 & 2593793 \\
			RSBench-2 & 3739494 & 3312846 & 3291100 \\
			RSBench-3 & 5017110 & 4994992 & 5106746 \\
			RSBench-4 & 2909850 & 2236588 & 3442009 \\
			RSBench-5 & 3901471 & 2916929 & 3574600 \\
			RSBench-6 & 6120710 & 4377109 & 5981138 \\
			RSBench-7 & 2814915 & 2435063 & 2715296 \\
			RSBench-8 & 3303001 & 2683367 & 2989405 \\
			RSBench-9 & 5279394 & 5148334 & 5192066 \\
			\hline
		\end{tabular}
	\end{table}
	
	{In the optimization process of RSBench, frequent calls to the LLM's API are required for both optimization and evaluation, leading to significant time and token consumption. In this subsection, we further analyze the time and token consumption during the optimization process.} Table~\ref{tab:time} presents the time each algorithm took to solve each problem. The results indicate that optimizing RSBench problems with LLM-based EAs is notably time-intensive, with each optimization run taking between approximately 2.5 and 8 hours. This high time consumption is primarily due to frequent interactions with the LLM for objective function evaluations in the RSBench problems. Both the inherent latency of LLM inference and the overhead from repeated communication contribute to the significant time costs associated with these evaluations. {Moreover, since the algorithms rely on calls to the GPT-3.5 API, the call time is influenced by the API provider. Therefore, it is more reasonable to compare the average number of tokens consumed across different algorithms. To facilitate this comparison, we also present the token consumption of the algorithms in Table~\ref{tab:token}.}
	
	\subsection{Effectiveness of the LLM-Based Initialization}
	\begin{figure}[!t]
		\begin{center}
			\subfigure[]{\label{fig:wo_init_train}\includegraphics[width=0.49\columnwidth]{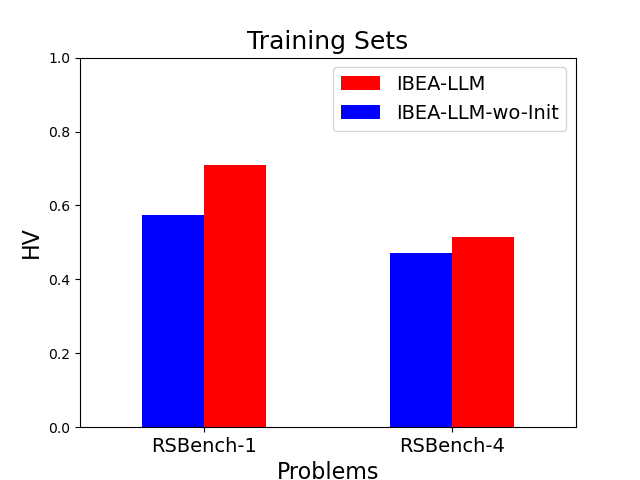}}
			\subfigure[]{\label{fig:wo_init_val}\includegraphics[width=0.49\columnwidth]{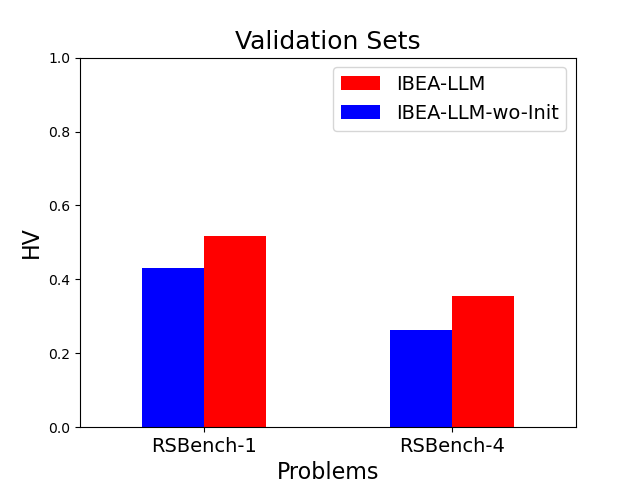}}
			\subfigure[]{\label{fig:wo_init_train_pf_1}\includegraphics[width=0.49\columnwidth]{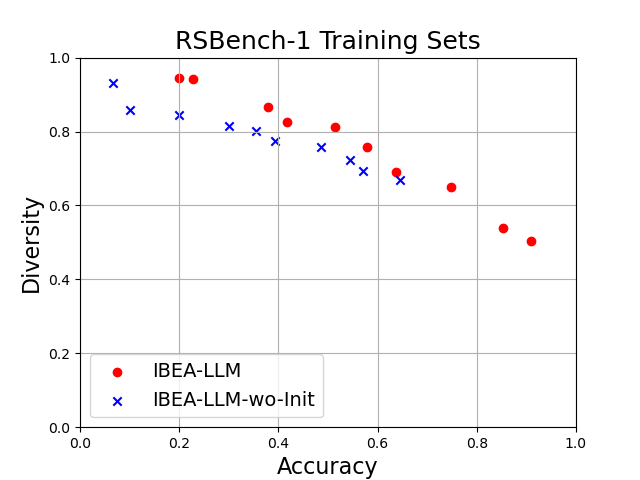}}
			\subfigure[]{\label{fig:wo_init_val_pf_1}\includegraphics[width=0.49\columnwidth]{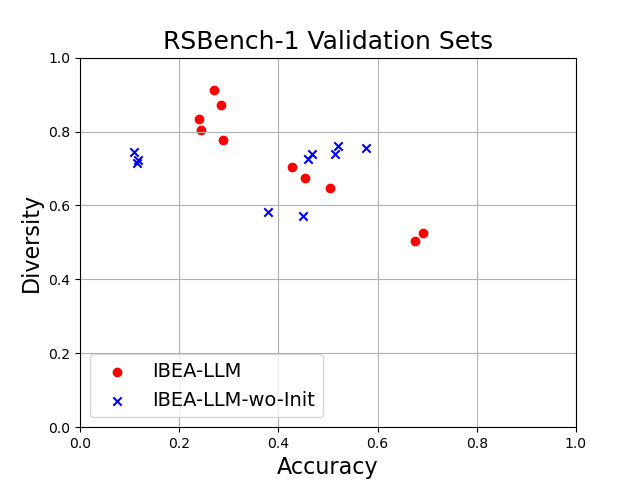}}
			\subfigure[]{\label{fig:wo_init_train_pf_4}\includegraphics[width=0.49\columnwidth]{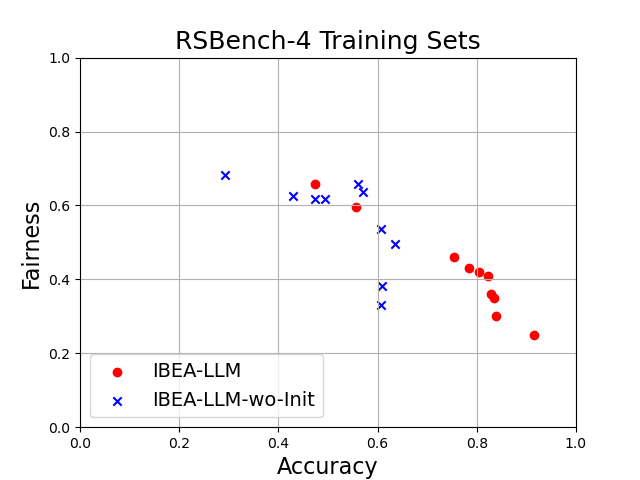}}
			\subfigure[]{\label{fig:wo_init_val_pf_4}\includegraphics[width=0.49\columnwidth]{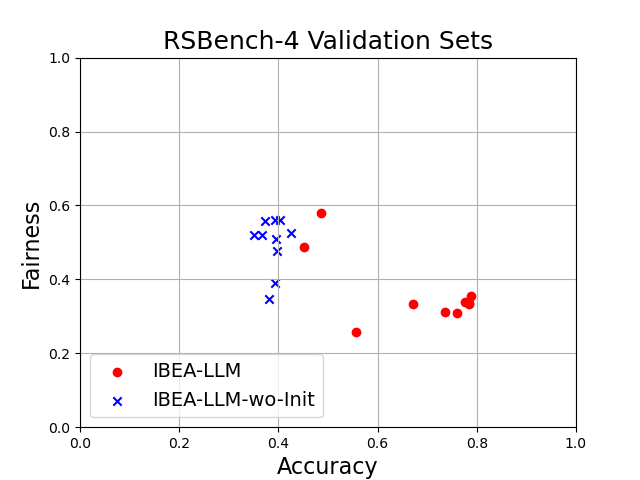}}
			\caption{Comparisons between LLM-IBEA and LLM-IBEA-wo-Init results. (a) HV results on RSBench-1 and RSBench-4 with the training sets. (b) HV results on RSBench-1 and RSBench-4 with the validation sets. (c) Pareto front approximations on RSBench-1 with the training set. (d) Pareto front approximations on RSBench-1 with the validation set. (e) Pareto front approximations on RSBench-4 with the training set. (f) Pareto front approximations on RSBench-4 with the validation set.}
			\label{fig:wo_init}
		\end{center}
	\end{figure}
	
	In Section~\ref{sec:initialization}, we introduce an initialization strategy designed to generate the initial population of recommendation prompts for the three LLM-based EAs considered in this study. This strategy involves providing an example chain-of-thought recommendation prompt, which the LLM uses as a basis to generate the initial population. To assess whether this initialization strategy improves the performance of LLM-based EAs, we examine its impact using LLM-IBEA, the best-performing algorithm. Specifically, we create a variant, LLM-IBEA-wo-Init, which omits the example prompt during initialization, enabling a direct comparison of the two approaches. RSBench-1 and RSBench-4 are used as representative problems in this investigation.
	
	Figure~\ref{fig:wo_init} presents the results of LLM-IBEA and its variant, LLM-IBEA-wo-Init, on the RSBench-1 and RSBench-4 problems. On the training set, as shown in Fig.~\ref{fig:wo_init_train}, LLM-IBEA achieves higher HV results compared to LLM-IBEA-wo-Init for both problems. The approximating points in the objective space, depicted in Fig.\ref{fig:wo_init_train_pf_1} and Fig.~\ref{fig:wo_init_train_pf_4}, reveal that LLM-IBEA demonstrates superior convergence. Additionally, the solutions provided by LLM-IBEA often include higher-accuracy recommendations, while LLM-IBEA-wo-Init lacks such high-accuracy solutions. On the validation set, similar trends are observed in Fig.\ref{fig:wo_init_val}, Fig.\ref{fig:wo_init_val_pf_1}, and Fig.~\ref{fig:wo_init_val_pf_4}, where LLM-IBEA continues to outperform LLM-IBEA-wo-Init. This performance difference can be explained by the chain-of-thought example used in the initialization strategy, which exhibits strong recommendation accuracy and is closer to the PF of the optimization problem. By adapting the recommendation prompt based on this example, the initialization strategy generates an initial population that is nearer to the PF and includes recommendation prompts with high accuracy, thereby supporting faster convergence and improved solution quality. 
	
	\subsection{Effectiveness of the LLM-Based Offspring Generation}
	\begin{figure}[!t]
		\begin{center}
			\subfigure[]{\label{fig:wo_cross_train}\includegraphics[width=0.49\columnwidth]{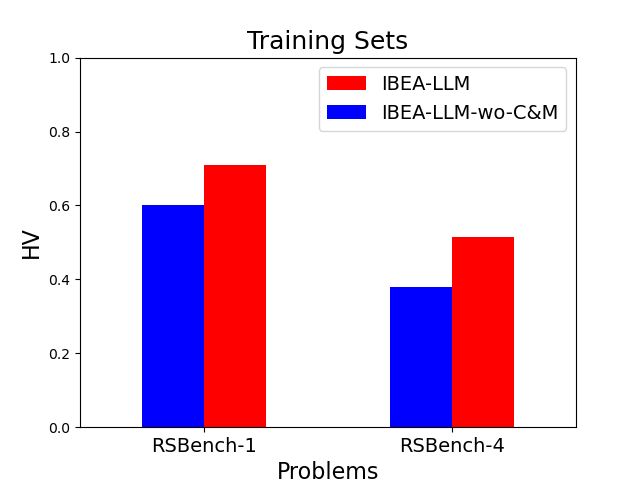}}
			\subfigure[]{\label{fig:wo_cross_val}\includegraphics[width=0.49\columnwidth]{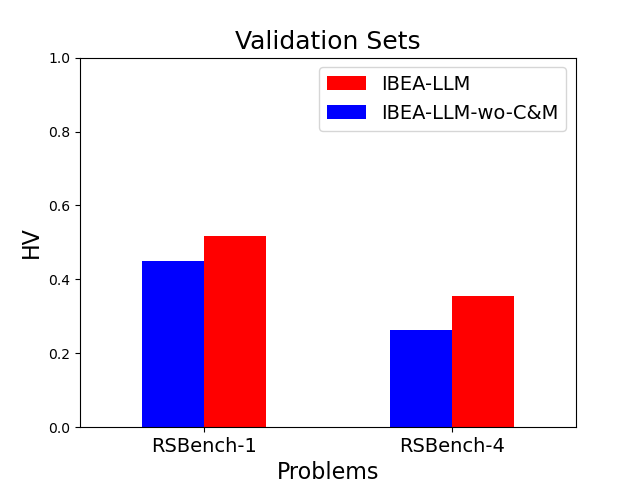}}
			\subfigure[]{\label{fig:wo_cross_train_pf_1}\includegraphics[width=0.49\columnwidth]{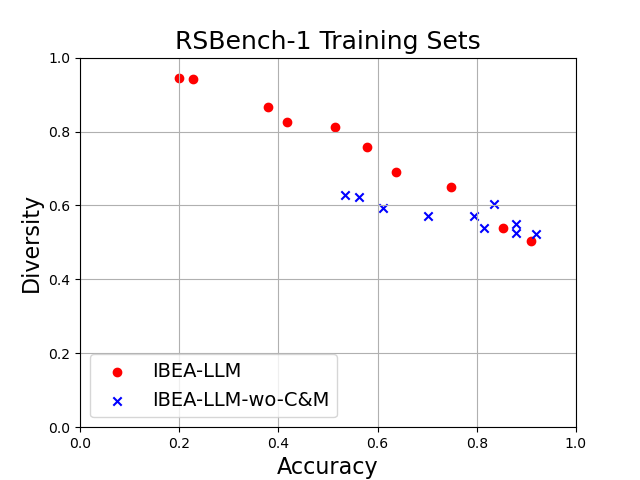}}
			\subfigure[]{\label{fig:wo_cross_val_pf_1}\includegraphics[width=0.49\columnwidth]{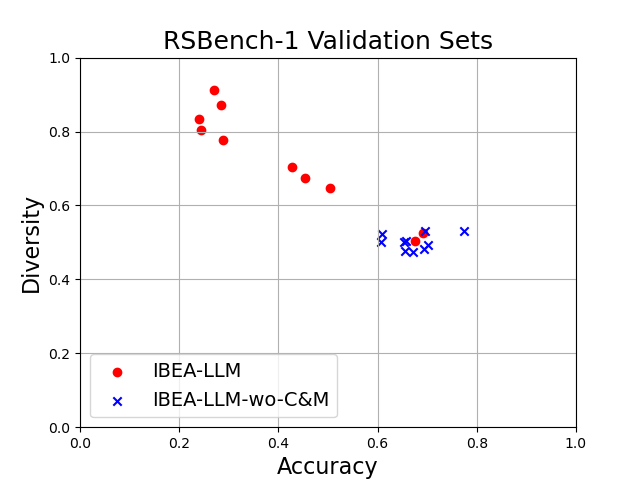}}
			\subfigure[]{\label{fig:wo_cross_train_pf_4}\includegraphics[width=0.49\columnwidth]{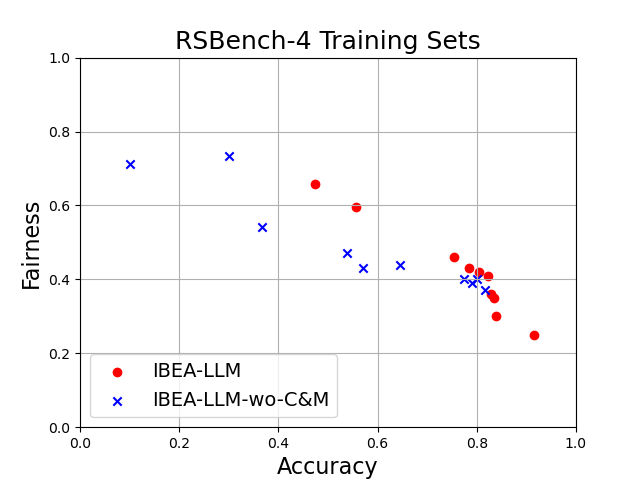}}
			\subfigure[]{\label{fig:wo_cross_val_pf_4}\includegraphics[width=0.49\columnwidth]{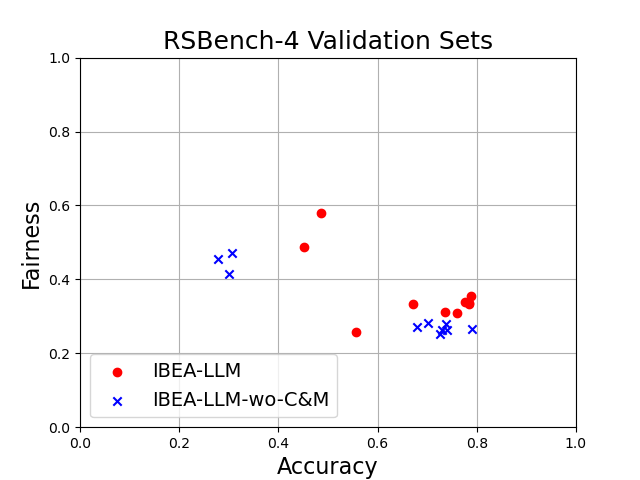}}
			\caption{Comparisons between LLM-IBEA and LLM-IBEA-wo-C\&M results. (a) HV results on RSBench-1 and RSBench-4 with the training sets. (b) HV results on RSBench-1 and RSBench-4 with the validation sets. (c) Pareto front approximations on RSBench-1 with the training set. (d) Pareto front approximations on RSBench-1 with the validation set. (e) Pareto front approximations on RSBench-4 with the training set. (f) Pareto front approximations on RSBench-4 with the validation set.}
			\label{fig:wo_cross}
		\end{center}
	\end{figure}
	
	As described in Section~\ref{sec:crossover}, the LLM-based EAs in this paper generate offspring recommendation prompts primarily through LLM-driven crossover and mutation operators. An important question to investigate is whether these LLM-driven crossover and mutation operators truly enhance algorithm performance on RSBench problems. To address this, we examine the performance of LLM-IBEA by comparing it with a variant named LLM-IBEA-wo-C\&M, which omits the LLM-driven crossover and mutation operations. In LLM-IBEA-wo-C\&M, offspring are generated directly based on all recommendation prompts stored in the population. The generation process is guided by the prompt: ‘\textit{Please generate a new prompt based on the following set: {\{population\}}. Provide a final prompt bracketed with $<$START$>$ and $<$END$>$.}’ We then evaluate the performance of both algorithms on RSBench-1 and RSBench-4.
	
	Fig.~\ref{fig:wo_cross} presents the results of LLM-IBEA and LLM-IBEA-wo-C\&M on RSBench-1 and RSBench-4. For the training sets, as shown in Fig.~\ref{fig:wo_cross_train}, LLM-IBEA outperforms LLM-IBEA-wo-C\&M on both problems in terms of HV results. Additionally, the approximated points in the objective space for LLM-IBEA tend to exhibit greater dispersion (i.e., Fig.\ref{fig:wo_cross_train_pf_1}) and improved convergence (i.e., Fig.\ref{fig:wo_cross_train_pf_4}), while the results from LLM-IBEA-wo-C\&M show limited diversity and suboptimal convergence. On the validation set, similar phenomenon can be observed from Fig.~\ref{fig:wo_cross_val}, Fig.~\ref{fig:wo_cross_val_pf_1}, and Fig.~\ref{fig:wo_cross_val_pf_4}. The results from these experiments suggest that the LLM-driven crossover and mutation operations enhance solution diversity in the objective space. This improvement may stem from the LLM-driven crossover and mutation, which, unlike direct LLM-based conditional generation, allow the algorithm to explore a broader search space, leading to improved exploration capabilities and greater diversity among the resulting solutions.
	
	{
		\subsection{Influence of the Neighborhood Size of LLM-MOEA/D}
		
		\begin{table}[!t]
			\centering
			\caption{Results of LLM-MOEA/D with the Neighborhood Size of 3, 5, and 7.}\label{tab:moead_param}
			\begin{tabular}{c|c|c|ccccccccc}
				\hline
				\multirow{2}{*}{Problems} & \multicolumn{3}{c}{LLM-MOEA/D with Different Neighborhood Sizes} \\
				\cline{2-4}
				& 0.3 & 0.5 & 0.7 \\
				\hline
				RSBench-1 & \textbf{0.6598$\pm$0.0633} & 0.6549$\pm$0.0187 & 0.6538$\pm$0.0339 \\
				RSBench-2 & \textbf{0.4215$\pm$0.0281} & 0.4131$\pm$0.0451 & 0.3734$\pm$0.0229 \\
				\hline
			\end{tabular}
		\end{table}
		
		In this subsection, we analyze the impact of neighborhood size on the performance of LLM-MOEA/D. Specifically, we evaluate the algorithm with neighborhood sizes of 3, 5, and 7 on RSBench-1 and RSBench-2. The results indicate that, for both RSBench-1 and RSBench-2, a neighborhood size of 3 yields the best performance. Therefore, we adopt this setting for the remaining experiments.
	}
	
	\section{What Can We Further Do Based on RSBench?}
	RSBench as a benchmark for RSs, providing strong evidence for the potential of EAs in LLM-based RSs. Looking ahead, RSBench is anticipated to catalyze further advancements in applying EAs within this field. In this section, we further present potential research points based on RSBench, with the goal of inspiring scholars in related disciplines to explore this promising area.
	
	\subsection{More Effective Objective Estimations}
	In RSBench, all objective function values are estimated using the Monte Carlo method, based on samples drawn from the training set. In our experimental studies, we randomly sample a batch of 10 samples from the training set to estimate each objective function value. However, it is clear that 10 samples may not be sufficient, and estimates based on this sample size may not accurately capture the quality of recommendation prompts. This limitation is reflected in our results, where the performance of optimized recommendation prompts on the test set is notably lower than on the training set. Increasing the sample size could mitigate this issue, but it would require more interactions with the LLM, significantly increasing the time and the token spent on evaluating each objective function. Consequently, designing more efficient, accurate, and time-saving objective function estimation methods is an area worth further exploration. Fortunately, RSBench can serve as a valuable platform for testing and evaluating these estimation techniques. 
	
	\subsection{More Efficient Evolutionary Algorithms}
	In the search for optimal recommendation prompts, EAs rely on iterative, trial-and-error processes, requiring numerous objective function evaluations to identify high-quality prompts. In our experiments, we set the iteration count to 20 and the population size to 10, yielding a total of 200 evaluations. Each evaluation involves interaction with the LLM, resulting in considerable time costs. Reducing evaluation costs during the evolutionary process is therefore a key research challenge. LLM-based surrogate models~\cite{hao2024large} or evolutionary transfer/multitasking techniques~\cite{7879282, 7161358} offer promising directions. RSBench thus serves as a valuable benchmark for evaluating the efficiency of these approaches.
	
	\subsection{More Powerful Recommended Capabilities}
	In this paper, the considered RS has a simple structure, relying on a single recommendation prompt to generate recommendations based on user-provided information. However, there is considerable potential to enhance both the design of recommendation prompts and the overall system architecture. For instance, EAs could be employed to generate more sophisticated prompts, such as chain-of-thought~\cite{wei2022chain} or even tree-of-thought~\cite{yao2024tree} prompts, to improve system performance. Additionally, integrating techniques like in-context learning~\cite{dong2022survey} or retrieval-augmented generation~\cite{gao2023retrieval}, with evolutionary computation optimizing relevant components, may make the LLM-based RS more robust. On such research topics, RSBench can serve as a flexible platform to support such expansions, enabling further exploration and innovation in this area.
	
	\section{Conclusion}
	
	To rigorously assess the performance and applicability of LLM-based EAs, real-world benchmark problems are essential. This paper primarily discusses the following three key points:
	\begin{enumerate}
		\item We introduce a benchmark problem set, RSBench, designed to evaluate LLM-based EAs in the context of recommendation prompt optimization, with a focus on session-based RS utilizing LLMs. RSBench aims to identify a set of Pareto-optimal prompts that guide the recommendation process, delivering recommendations that are accurate, diverse, and fair.
		
		\item We integrate an LLM-based initialization strategy with LLM-driven offspring generation strategies to develop three LLM-based EAs—LLM-NSGA-II, LLM-MOEA/D, and LLM-IBEA—within established multiobjective evolutionary frameworks. Performance evaluations on RSBench demonstrate that LLM-IBEA achieves the best results across the benchmark problems.
		
		\item We conduct experimental investigations to evaluate the effectiveness of LLM-based initialization and LLM-driven offspring generation in enhancing the overall performance of the algorithms.
	\end{enumerate}
	In addition to these three points, we also discuss potential future research directions that RSBench opens for LLM-based recommendation systems and EAs. We believe RSBench will significantly contribute to advancing LLM-based EAs and recommendation systems. 
	Further work will focus on developing more effective objective estimation methods to improve the performance of LLM-based EAs in RSs, designing more efficient EA frameworks to tackle complex optimization problems in RSs, and exploring the potential of LLM-based EAs in enabling more powerful recommended capabilities in RSs.

	
	%

	\appendices
	

	\ifCLASSOPTIONcaptionsoff
	\newpage
	\fi
	
	\bibliographystyle{IEEEtran}
	\bibliography{rsbench_bib}

\begin{thebibliography}{10}
\providecommand{\url}[1]{#1}
\csname url@samestyle\endcsname
\providecommand{\newblock}{\relax}
\providecommand{\bibinfo}[2]{#2}
\providecommand{\BIBentrySTDinterwordspacing}{\spaceskip=0pt\relax}
\providecommand{\BIBentryALTinterwordstretchfactor}{4}
\providecommand{\BIBentryALTinterwordspacing}{\spaceskip=\fontdimen2\font plus
\BIBentryALTinterwordstretchfactor\fontdimen3\font minus
  \fontdimen4\font\relax}
\providecommand{\BIBforeignlanguage}[2]{{%
\expandafter\ifx\csname l@#1\endcsname\relax
\typeout{** WARNING: IEEEtran.bst: No hyphenation pattern has been}%
\typeout{** loaded for the language `#1'. Using the pattern for}%
\typeout{** the default language instead.}%
\else
\language=\csname l@#1\endcsname
\fi
#2}}
\providecommand{\BIBdecl}{\relax}
\BIBdecl

\bibitem{coello2007evolutionary}
C.~A.~C. Coello, \emph{Evolutionary algorithms for solving multi-objective
  problems}.\hskip 1em plus 0.5em minus 0.4em\relax Springer, 2007.

\bibitem{oganov2019structure}
A.~R. Oganov, C.~J. Pickard, Q.~Zhu, and R.~J. Needs, ``Structure prediction
  drives materials discovery,'' \emph{Nature Reviews Materials}, vol.~4, no.~5,
  pp. 331--348, 2019.

\bibitem{deb1999evolutionary}
K.~Deb \emph{et~al.}, ``Evolutionary algorithms for multi-criterion
  optimization in engineering design,'' \emph{Evolutionary algorithms in
  engineering and computer science}, vol.~2, pp. 135--161, 1999.

\bibitem{branke2008multiobjective}
J.~Branke, \emph{Multiobjective optimization: Interactive and evolutionary
  approaches}.\hskip 1em plus 0.5em minus 0.4em\relax Springer Science \&
  Business Media, 2008, vol. 5252.

\bibitem{9005525}
A.~Gupta, Y.-S. Ong, M.~Shakeri, X.~Chi, and A.~Z. NengSheng, ``The blessing of
  dimensionality in many-objective search: An inverse machine learning
  insight,'' in \emph{2019 IEEE International Conference on Big Data (Big
  Data)}, 2019, pp. 3896--3902.

\bibitem{zhao2023survey}
W.~X. Zhao, K.~Zhou, J.~Li, T.~Tang, X.~Wang, Y.~Hou, Y.~Min, B.~Zhang,
  J.~Zhang, Z.~Dong \emph{et~al.}, ``A survey of large language models,''
  \emph{arXiv preprint arXiv:2303.18223}, 2023.

\bibitem{OpenAI}
\BIBentryALTinterwordspacing
OpenAI, ``Chatgpt 3.5.'' [Online]. Available:
  \url{https://platform.openai.com/docs/api-reference}
\BIBentrySTDinterwordspacing

\bibitem{buehler2024mechgpt}
M.~J. Buehler, ``Mech{GPT}, a language-based strategy for mechanics and
  materials modeling that connects knowledge across scales, disciplines, and
  modalities,'' \emph{Applied Mechanics Reviews}, vol.~76, no.~2, p. 021001,
  2024.

\bibitem{li2023large}
B.~Li, K.~Mellou, B.~Zhang, J.~Pathuri, and I.~Menache, ``Large language models
  for supply chain optimization,'' \emph{arXiv preprint arXiv:2307.03875},
  2023.

\bibitem{bran2023chemcrow}
A.~M. Bran, S.~Cox, O.~Schilter, C.~Baldassari, A.~D. White, and P.~Schwaller,
  ``Chem{C}row: Augmenting large-language models with chemistry tools,''
  \emph{arXiv preprint arXiv:2304.05376}, 2023.

\bibitem{wu2024evolutionary}
X.~Wu, S.-h. Wu, J.~Wu, L.~Feng, and K.~C. Tan, ``Evolutionary computation in
  the era of large language model: Survey and roadmap,'' \emph{arXiv preprint
  arXiv:2401.10034}, 2024.

\bibitem{guo2023connecting}
Q.~Guo, R.~Wang, J.~Guo, B.~Li, K.~Song, X.~Tan, G.~Liu, J.~Bian, and Y.~Yang,
  ``Connecting large language models with evolutionary algorithms yields
  powerful prompt optimizers,'' \emph{arXiv preprint arXiv:2309.08532}, 2023.

\bibitem{10.1145/3694791}
\BIBentryALTinterwordspacing
E.~Meyerson, M.~J. Nelson, H.~Bradley, A.~Gaier, A.~Moradi, A.~K. Hoover, and
  J.~Lehman, ``Language model crossover: Variation through few-shot
  prompting,'' \emph{ACM Trans. Evol. Learn. Optim.}, Sep. 2024, just Accepted.
  [Online]. Available: \url{https://doi.org/10.1145/3694791}
\BIBentrySTDinterwordspacing

\bibitem{xu2022gps}
H.~Xu, Y.~Chen, Y.~Du, N.~Shao, Y.~Wang, H.~Li, and Z.~Yang, ``{GPS}: Genetic
  prompt search for efficient few-shot learning,'' \emph{arXiv preprint
  arXiv:2210.17041}, 2022.

\bibitem{nasir2024llmatic}
M.~U. Nasir, S.~Earle, J.~Togelius, S.~James, and C.~Cleghorn, ``{LLM}atic:
  neural architecture search via large language models and quality diversity
  optimization,'' in \emph{Proceedings of the Genetic and Evolutionary
  Computation Conference}, 2024, pp. 1110--1118.

\bibitem{bradley2024openelm}
H.~Bradley, H.~Fan, T.~Galanos, R.~Zhou, D.~Scott, and J.~Lehman, ``The openelm
  library: Leveraging progress in language models for novel evolutionary
  algorithms,'' in \emph{Genetic Programming Theory and Practice XX}.\hskip 1em
  plus 0.5em minus 0.4em\relax Springer, 2024, pp. 177--201.

\bibitem{bobadilla2013recommender}
J.~Bobadilla, F.~Ortega, A.~Hernando, and A.~Guti{\'e}rrez, ``Recommender
  systems survey,'' \emph{Knowledge-based systems}, vol.~46, pp. 109--132,
  2013.

\bibitem{10506571}
Z.~Zhao, W.~Fan, J.~Li, Y.~Liu, X.~Mei, Y.~Wang, Z.~Wen, F.~Wang, X.~Zhao,
  J.~Tang, and Q.~Li, ``Recommender systems in the era of large language models
  (llms),'' \emph{IEEE Transactions on Knowledge and Data Engineering},
  vol.~36, no.~11, pp. 6889--6907, 2024.

\bibitem{koren2021advances}
Y.~Koren, S.~Rendle, and R.~Bell, ``Advances in collaborative filtering,''
  \emph{Recommender systems handbook}, pp. 91--142, 2021.

\bibitem{ren2024representation}
X.~Ren, W.~Wei, L.~Xia, L.~Su, S.~Cheng, J.~Wang, D.~Yin, and C.~Huang,
  ``Representation learning with large language models for recommendation,'' in
  \emph{Proceedings of the ACM on Web Conference 2024}, 2024, pp. 3464--3475.

\bibitem{ji2024genrec}
J.~Ji, Z.~Li, S.~Xu, W.~Hua, Y.~Ge, J.~Tan, and Y.~Zhang, ``Genrec: Large
  language model for generative recommendation,'' in \emph{European Conference
  on Information Retrieval}.\hskip 1em plus 0.5em minus 0.4em\relax Springer,
  2024, pp. 494--502.

\bibitem{10.1145/3626772.3657688}
\BIBentryALTinterwordspacing
Z.~Sun, H.~Liu, X.~Qu, K.~Feng, Y.~Wang, and Y.-S. Ong, ``Large language models
  for intent-driven session recommendations,'' in \emph{Proceedings of the 47th
  International ACM SIGIR Conference on Research and Development in Information
  Retrieval}, ser. SIGIR '24.\hskip 1em plus 0.5em minus 0.4em\relax New York,
  NY, USA: Association for Computing Machinery, 2024, p. 324–334. [Online].
  Available: \url{https://doi-org.remotexs.ntu.edu.sg/10.1145/3626772.3657688}
\BIBentrySTDinterwordspacing

\bibitem{kenton2019bert}
J.~D. M.-W.~C. Kenton and L.~K. Toutanova, ``{BERT}: Pre-training of deep
  bidirectional transformers for language understanding,'' in \emph{Proceedings
  of naacL-HLT}, vol.~1.\hskip 1em plus 0.5em minus 0.4em\relax Minneapolis,
  Minnesota, 2019, p.~2.

\bibitem{hoang2021using}
B.~N.~M. Hoang, H.~T.~H. Vy, T.~G. Hong, V.~T.~M. Hang, H.~L. T.~K. Nhung
  \emph{et~al.}, ``Using {BERT} embedding to improve memory-based collaborative
  filtering recommender systems,'' in \emph{2021 RIVF International Conference
  on Computing and Communication Technologies (RIVF)}.\hskip 1em plus 0.5em
  minus 0.4em\relax IEEE, 2021, pp. 1--6.

\bibitem{wu2020ptum}
C.~Wu, F.~Wu, T.~Qi, J.~Lian, Y.~Huang, and X.~Xie, ``{PTUM}: Pre-training user
  model from unlabeled user behaviors via self-supervision,'' \emph{arXiv
  preprint arXiv:2010.01494}, 2020.

\bibitem{friedman2023leveraging}
L.~Friedman, S.~Ahuja, D.~Allen, Z.~Tan, H.~Sidahmed, C.~Long, J.~Xie,
  G.~Schubiner, A.~Patel, H.~Lara \emph{et~al.}, ``Leveraging large language
  models in conversational recommender systems,'' \emph{arXiv preprint
  arXiv:2305.07961}, 2023.

\bibitem{mao2023unitrec}
Z.~Mao, H.~Wang, Y.~Du, and K.-F. Wong, ``Unitrec: A unified text-to-text
  transformer and joint contrastive learning framework for text-based
  recommendation,'' \emph{arXiv preprint arXiv:2305.15756}, 2023.

\bibitem{wang2023recmind}
Y.~Wang, Z.~Jiang, Z.~Chen, F.~Yang, Y.~Zhou, E.~Cho, X.~Fan, X.~Huang, Y.~Lu,
  and Y.~Yang, ``Recmind: Large language model powered agent for
  recommendation,'' \emph{arXiv preprint arXiv:2308.14296}, 2023.

\bibitem{huang2023recommender}
X.~Huang, J.~Lian, Y.~Lei, J.~Yao, D.~Lian, and X.~Xie, ``Recommender {AI}
  agent: Integrating large language models for interactive recommendations,''
  \emph{arXiv preprint arXiv:2308.16505}, 2023.

\bibitem{liu2023chatgpt}
J.~Liu, C.~Liu, P.~Zhou, R.~Lv, K.~Zhou, and Y.~Zhang, ``Is {ChatGPT} a good
  recommender? a preliminary study,'' \emph{arXiv preprint arXiv:2304.10149},
  2023.

\bibitem{zhiyuli2023bookgpt}
A.~Zhiyuli, Y.~Chen, X.~Zhang, and X.~Liang, ``Book{GPT}: A general framework
  for book recommendation empowered by large language model,'' \emph{arXiv
  preprint arXiv:2305.15673}, 2023.

\bibitem{zhang2008avoiding}
M.~Zhang and N.~Hurley, ``Avoiding monotony: improving the diversity of
  recommendation lists,'' in \emph{Proceedings of the 2008 ACM conference on
  Recommender systems}, 2008, pp. 123--130.

\bibitem{li2022fairness}
Y.~Li, H.~Chen, S.~Xu, Y.~Ge, J.~Tan, S.~Liu, and Y.~Zhang, ``Fairness in
  recommendation: A survey,'' \emph{arXiv preprint arXiv:2205.13619}, 2022.

\bibitem{rodriguez2012multiple}
M.~Rodriguez, C.~Posse, and E.~Zhang, ``Multiple objective optimization in
  recommender systems,'' in \emph{Proceedings of the sixth ACM conference on
  Recommender systems}, 2012, pp. 11--18.

\bibitem{qian-good}
C.~Qian, ``{Multiobjective Evolutionary Algorithms Are Still Good: Maximizing
  Monotone Approximately Submodular Minus Modular Functions},''
  \emph{Evolutionary Computation}, vol.~29, no.~4, pp. 463--490, 12 2021.

\bibitem{nsga2}
K.~Deb, A.~Pratap, S.~Agarwal, and T.~Meyarivan, ``A fast and elitist
  multiobjective genetic algorithm: {NSGA}-{II},'' \emph{IEEE Trans. on Evol.
  Comput.}, vol.~6, no.~2, pp. 182--197, 2002.

\bibitem{spea2}
E.~Zitzler, M.~Laumanns, and L.~Thiele, ``{SPEA2}: Improving the strength
  pareto evolutionary algorithm,'' \emph{TIK-report}, vol. 103, 2001.

\bibitem{decomposition-survey}
A.~Trivedi, D.~Srinivasan, K.~Sanyal, and A.~Ghosh, ``A survey of
  multiobjective evolutionary algorithms based on decomposition,'' \emph{IEEE
  Transactions on Evolutionary Computation}, vol.~21, no.~3, pp. 440--462,
  2017.

\bibitem{moead}
Q.~Zhang and H.~Li, ``{MOEA/D}: A multiobjective evolutionary algorithm based
  on decomposition,'' \emph{IEEE Trans. on Evol. Comput.}, vol.~11, no.~6, pp.
  712--731, 2007.

\bibitem{fvmoea}
S.~{Jiang}, J.~{Zhang}, Y.~{Ong}, A.~N. {Zhang}, and P.~S. {Tan}, ``A simple
  and fast hypervolume indicator-based multiobjective evolutionary algorithm,''
  \emph{IEEE Trans. on Cybern.}, vol.~45, no.~10, pp. 2202--2213, 2015.

\bibitem{ibea}
E.~Zitzler and S.~K{\"u}nzli, ``Indicator-based selection in multiobjective
  search,'' in \emph{Parallel Problem Solving from Nat. - PPSN VIII}, X.~Yao,
  E.~K. Burke, J.~A. Lozano, J.~Smith, J.~J. Merelo-Guerv{\'o}s, J.~A.
  Bullinaria, J.~E. Rowe, P.~Ti{\v{n}}o, A.~Kab{\'a}n, and H.-P. Schwefel,
  Eds.\hskip 1em plus 0.5em minus 0.4em\relax Berlin, Heidelberg: Springer
  Berlin Heidelberg, 2004, pp. 832--842.

\bibitem{yang2013grid}
S.~Yang, M.~Li, X.~Liu, and J.~Zheng, ``A grid-based evolutionary algorithm for
  many-objective optimization,'' \emph{IEEE Transactions on Evolutionary
  Computation}, vol.~17, no.~5, pp. 721--736, 2013.

\bibitem{8632683}
Z.~Ma and Y.~Wang, ``Evolutionary constrained multiobjective optimization: Test
  suite construction and performance comparisons,'' \emph{IEEE Transactions on
  Evolutionary Computation}, vol.~23, no.~6, pp. 972--986, 2019.

\bibitem{8624421}
Z.-Z. Liu and Y.~Wang, ``Handling constrained multiobjective optimization
  problems with constraints in both the decision and objective spaces,''
  \emph{IEEE Transactions on Evolutionary Computation}, vol.~23, no.~5, pp.
  870--884, 2019.

\bibitem{shao2021multi}
Y.~Shao, J.~C.-W. Lin, G.~Srivastava, D.~Guo, H.~Zhang, H.~Yi, and A.~Jolfaei,
  ``Multi-objective neural evolutionary algorithm for combinatorial
  optimization problems,'' \emph{IEEE transactions on neural networks and
  learning systems}, vol.~34, no.~4, pp. 2133--2143, 2021.

\bibitem{liu2024large}
S.~Liu, C.~Chen, X.~Qu, K.~Tang, and Y.-S. Ong, ``Large language models as
  evolutionary optimizers,'' in \emph{2024 IEEE Congress on Evolutionary
  Computation (CEC)}.\hskip 1em plus 0.5em minus 0.4em\relax IEEE, 2024, pp.
  1--8.

\bibitem{yang2024largelanguagemodelsoptimizers}
\BIBentryALTinterwordspacing
C.~Yang, X.~Wang, Y.~Lu, H.~Liu, Q.~V. Le, D.~Zhou, and X.~Chen, ``Large
  language models as optimizers,'' 2024. [Online]. Available:
  \url{https://arxiv.org/abs/2309.03409}
\BIBentrySTDinterwordspacing

\bibitem{hao2024large}
H.~Hao, X.~Zhang, and A.~Zhou, ``Large language models as surrogate models in
  evolutionary algorithms: A preliminary study,'' \emph{arXiv preprint
  arXiv:2406.10675}, 2024.

\bibitem{wang2024large}
Z.~Wang, S.~Liu, J.~Chen, and K.~C. Tan, ``Large language model-aided
  evolutionary search for constrained multiobjective optimization,'' in
  \emph{International Conference on Intelligent Computing}.\hskip 1em plus
  0.5em minus 0.4em\relax Springer, 2024, pp. 218--230.

\bibitem{liu2023large}
F.~Liu, X.~Lin, Z.~Wang, S.~Yao, X.~Tong, M.~Yuan, and Q.~Zhang, ``Large
  language model for multi-objective evolutionary optimization,'' \emph{arXiv
  preprint arXiv:2310.12541}, 2023.

\bibitem{wong2024llm2fea}
M.~Wong, J.~Liu, T.~Rios, S.~Menzel, and Y.-S. Ong, ``{LLM2FEA}: Discover novel
  designs with generative evolutionary multitasking,'' \emph{arXiv preprint
  arXiv:2406.14917}, 2024.

\bibitem{hemberg2024evolving}
E.~Hemberg, S.~Moskal, and U.-M. O’Reilly, ``Evolving code with a large
  language model,'' \emph{Genetic Programming and Evolvable Machines}, vol.~25,
  no.~2, p.~21, 2024.

\bibitem{brownlee2023enhancing}
A.~E. Brownlee, J.~Callan, K.~Even-Mendoza, A.~Geiger, C.~Hanna, J.~Petke,
  F.~Sarro, and D.~Sobania, ``Enhancing genetic improvement mutations using
  large language models,'' in \emph{International Symposium on Search Based
  Software Engineering}.\hskip 1em plus 0.5em minus 0.4em\relax Springer, 2023,
  pp. 153--159.

\bibitem{yu2023gpt}
C.~Yu, X.~Liu, Y.~Wang, Y.~Liu, W.~Feng, X.~Deng, C.~Tang, and J.~Lv,
  ``{GPT-NAS}: Evolutionary neural architecture search with the generative
  pre-trained model,'' \emph{arXiv preprint arXiv:2305.05351}, 2023.

\bibitem{diao2022black}
S.~Diao, Z.~Huang, R.~Xu, X.~Li, Y.~Lin, X.~Zhou, and T.~Zhang, ``Black-box
  prompt learning for pre-trained language models,'' \emph{arXiv preprint
  arXiv:2201.08531}, 2022.

\bibitem{saletta2024exploring}
M.~Saletta and C.~Ferretti, ``Exploring the prompt space of large language
  models through evolutionary sampling,'' in \emph{Proceedings of the Genetic
  and Evolutionary Computation Conference}, 2024, pp. 1345--1353.

\bibitem{xu2023wizardlm}
C.~Xu, Q.~Sun, K.~Zheng, X.~Geng, P.~Zhao, J.~Feng, C.~Tao, and D.~Jiang,
  ``Wizardlm: Empowering large language models to follow complex
  instructions,'' \emph{arXiv preprint arXiv:2304.12244}, 2023.

\bibitem{li2022competition}
Y.~Li, D.~Choi, J.~Chung, N.~Kushman, J.~Schrittwieser, R.~Leblond, T.~Eccles,
  J.~Keeling, F.~Gimeno, A.~Dal~Lago \emph{et~al.}, ``Competition-level code
  generation with alphacode,'' \emph{Science}, vol. 378, no. 6624, pp.
  1092--1097, 2022.

\bibitem{wu2023deceptprompt}
F.~Wu, X.~Liu, and C.~Xiao, ``Deceptprompt: Exploiting {LLM}-driven code
  generation via adversarial natural language instructions,'' \emph{arXiv
  preprint arXiv:2312.04730}, 2023.

\bibitem{morris2024llm}
C.~Morris, M.~Jurado, and J.~Zutty, ``L{LM} guided evolution-the automation of
  models advancing models,'' in \emph{Proceedings of the Genetic and
  Evolutionary Computation Conference}, 2024, pp. 377--384.

\bibitem{geng2015nnia}
B.~Geng, L.~Li, L.~Jiao, M.~Gong, Q.~Cai, and Y.~Wu, ``{NNIA-RS}: A
  multi-objective optimization based recommender system,'' \emph{Physica A:
  Statistical Mechanics and its Applications}, vol. 424, pp. 383--397, 2015.

\bibitem{cui2017novel}
L.~Cui, P.~Ou, X.~Fu, Z.~Wen, and N.~Lu, ``A novel multi-objective evolutionary
  algorithm for recommendation systems,'' \emph{Journal of Parallel and
  Distributed Computing}, vol. 103, pp. 53--63, 2017.

\bibitem{cai2020hybrid}
X.~Cai, Z.~Hu, P.~Zhao, W.~Zhang, and J.~Chen, ``A hybrid recommendation system
  with many-objective evolutionary algorithm,'' \emph{Expert Systems with
  Applications}, vol. 159, p. 113648, 2020.

\bibitem{rana2014evolutionary}
C.~Rana and S.~K. Jain, ``An evolutionary clustering algorithm based on
  temporal features for dynamic recommender systems,'' \emph{Swarm and
  Evolutionary Computation}, vol.~14, pp. 21--30, 2014.

\bibitem{asgarnezhad2022effective}
R.~Asgarnezhad, S.~S. AbdullMajeed, Z.~A. Abbas, and S.~S. Salman, ``An
  effective algorithm to improve recommender systems using evolutionary
  computation algorithms and neural network,'' \emph{Wasit Journal of Computer
  and Mathematics Science}, vol.~1, no.~1, 2022.

\bibitem{milojkovic2019multi}
N.~Milojkovic, D.~Antognini, G.~Bergamin, B.~Faltings, and C.~Musat,
  ``Multi-gradient descent for multi-objective recommender systems,''
  \emph{arXiv preprint arXiv:2001.00846}, 2019.

\bibitem{ni2019justifying}
J.~Ni, J.~Li, and J.~McAuley, ``Justifying recommendations using
  distantly-labeled reviews and fine-grained aspects,'' in \emph{Proceedings of
  the 2019 conference on empirical methods in natural language processing and
  the 9th international joint conference on natural language processing
  (EMNLP-IJCNLP)}, 2019, pp. 188--197.

\bibitem{steck2011item}
H.~Steck, ``Item popularity and recommendation accuracy,'' in \emph{Proceedings
  of the fifth ACM conference on Recommender systems}, 2011, pp. 125--132.

\bibitem{kunaver2017diversity}
M.~Kunaver and T.~Po{\v{z}}rl, ``Diversity in recommender systems--a survey,''
  \emph{Knowledge-based systems}, vol. 123, pp. 154--162, 2017.

\bibitem{yin2023understanding}
Q.~Yin, H.~Fang, Z.~Sun, and Y.-S. Ong, ``Understanding diversity in
  session-based recommendation,'' \emph{ACM Transactions on Information
  Systems}, vol.~42, no.~1, pp. 1--34, 2023.

\bibitem{abdollahpouri2017controlling}
H.~Abdollahpouri, R.~Burke, and B.~Mobasher, ``Controlling popularity bias in
  learning-to-rank recommendation,'' in \emph{Proceedings of the eleventh ACM
  conference on recommender systems}, 2017, pp. 42--46.

\bibitem{sun2022revisiting}
Z.~Sun, J.~Yang, K.~Feng, H.~Fang, X.~Qu, and Y.-S. Ong, ``Revisiting bundle
  recommendation: Datasets, tasks, challenges and opportunities for
  intent-aware product bundling,'' in \emph{Proceedings of the 45th
  International ACM SIGIR Conference on Research and Development in Information
  Retrieval}, 2022, pp. 2900--2911.

\bibitem{wei2022chain}
J.~Wei, X.~Wang, D.~Schuurmans, M.~Bosma, F.~Xia, E.~Chi, Q.~V. Le, D.~Zhou
  \emph{et~al.}, ``Chain-of-thought prompting elicits reasoning in large
  language models,'' \emph{Advances in neural information processing systems},
  vol.~35, pp. 24\,824--24\,837, 2022.

\bibitem{yao2024tree}
S.~Yao, D.~Yu, J.~Zhao, I.~Shafran, T.~Griffiths, Y.~Cao, and K.~Narasimhan,
  ``Tree of thoughts: Deliberate problem solving with large language models,''
  \emph{Advances in Neural Information Processing Systems}, vol.~36, 2024.

\bibitem{ref_dirs_energy}
J.~{Blank}, K.~{Deb}, Y.~{Dhebar}, S.~{Bandaru}, and H.~{Seada}, ``Generating
  well-spaced points on a unit simplex for evolutionary many-objective
  optimization,'' \emph{IEEE Transactions on Evolutionary Computation}, In
  press.

\bibitem{auger2012hypervolume}
A.~Auger, J.~Bader, D.~Brockhoff, and E.~Zitzler, ``Hypervolume-based
  multiobjective optimization: Theoretical foundations and practical
  implications,'' \emph{Theoretical Computer Science}, vol. 425, pp. 75--103,
  2012.

\bibitem{7879282}
L.~Feng, Y.-S. Ong, S.~Jiang, and A.~Gupta, ``Autoencoding evolutionary search
  with learning across heterogeneous problems,'' \emph{IEEE Transactions on
  Evolutionary Computation}, vol.~21, no.~5, pp. 760--772, 2017.

\bibitem{7161358}
A.~Gupta, Y.-S. Ong, and L.~Feng, ``Multifactorial evolution: Toward
  evolutionary multitasking,'' \emph{IEEE Transactions on Evolutionary
  Computation}, vol.~20, no.~3, pp. 343--357, 2016.

\bibitem{dong2022survey}
Q.~Dong, L.~Li, D.~Dai, C.~Zheng, J.~Ma, R.~Li, H.~Xia, J.~Xu, Z.~Wu, T.~Liu
  \emph{et~al.}, ``A survey on in-context learning,'' \emph{arXiv preprint
  arXiv:2301.00234}, 2022.

\bibitem{gao2023retrieval}
Y.~Gao, Y.~Xiong, X.~Gao, K.~Jia, J.~Pan, Y.~Bi, Y.~Dai, J.~Sun, M.~Wang, and
  H.~Wang, ``Retrieval-augmented generation for large language models: A
  survey,'' \emph{arXiv preprint arXiv:2312.10997}, 2023.

\end{thebibliography}

\end{document}